\documentclass[sigconf, nonacm]{acmart}

\AtBeginDocument{%
  }

\usepackage{graphicx}
\usepackage{multicol,multirow}
\usepackage[figuresright]{rotating}
\usepackage{appendix}
\usepackage{ifpdf}
\usepackage[T1]{fontenc}
\usepackage{textcomp}
\usepackage{xcolor}
\usepackage{float} 
\usepackage{booktabs} 
\usepackage{enumitem} 
\usepackage{multicol,multirow}
\usepackage{array}
\usepackage{makecell}
\usepackage{longtable}
\usepackage{colortbl}
\usepackage[bottom]{footmisc}
\usepackage{url}
\usepackage{graphics}
\usepackage{subcaption}
\usepackage{grffile} 
\usepackage{tabularray}
\usepackage{ctable} 
\usepackage{makecell} 
\usepackage{svg}
\usepackage{indentfirst}
\usepackage{mathtools}
\usepackage{setspace}
\usepackage{wrapfig}
\usepackage{lscape}
\usepackage{rotating}
\usepackage{epstopdf}
\usepackage{relsize}
\usepackage{microtype} 
\usepackage{ragged2e}

\emergencystretch=\maxdimen
\begin{document}

\relscale{1.0}


\title{Progressive Ideation using an Agentic AI Framework for Human-AI Co-Creation}

\author{B. Sankar}
\authornote{Corresponding author}
\affiliation{%
  \institution{Department of Mechanical Engineering,\\ Indian Institute of Science (IISc),}
  \city{Bangalore}
  \postcode{560012}
  \country{India}
}
\email{sankarb@iisc.ac.in}
\orcid{}

\author{Srinidhi Ranjini Girish}
\affiliation{%
  \institution{Department of Design and Manufacturing,\\ Indian Institute of Science (IISc),}
  \city{Bangalore}
  \postcode{560012}
  \country{India}
}
\orcid{}

\author{Aadya Bharti}
\affiliation{%
  \institution{Department of Design and Manufacturing,\\ Indian Institute of Science (IISc),}
  \city{Bangalore}
  \postcode{560012}
  \country{India}
}
\orcid{}

\author{Dibakar Sen}
\affiliation{%
  \institution{Department of Design and Manufacturing,\\ Indian Institute of Science (IISc),}
  \city{Bangalore}
  \postcode{560012}
  \country{India}
}
\orcid{}

\renewcommand{\shortauthors}{B. Sankar et al.}


\begin{abstract}
The generation of truly novel and diverse ideas is important for contemporary engineering design, yet it remains a significant cognitive challenge for novice designers. Current 'single-spurt' AI systems exacerbate this challenge by producing a high volume of semantically clustered ideas. We propose MIDAS (Meta-cognitive Ideation through Distributed Agentic AI System), a novel framework that replaces the single-AI paradigm with a distributed 'team' of specialized AI agents designed to emulate the human meta-cognitive ideation workflow. This agentic system progressively refines ideas and assesses each one for both global novelty (against existing solutions) and local novelty (against previously generated ideas). MIDAS, therefore, demonstrates a viable and progressive paradigm for true human-AI co-creation, elevating the human designer from a passive filterer to a participatory, active, collaborative partner.
\end{abstract}



\begin{teaserfigure}
  \includegraphics[width=\linewidth]{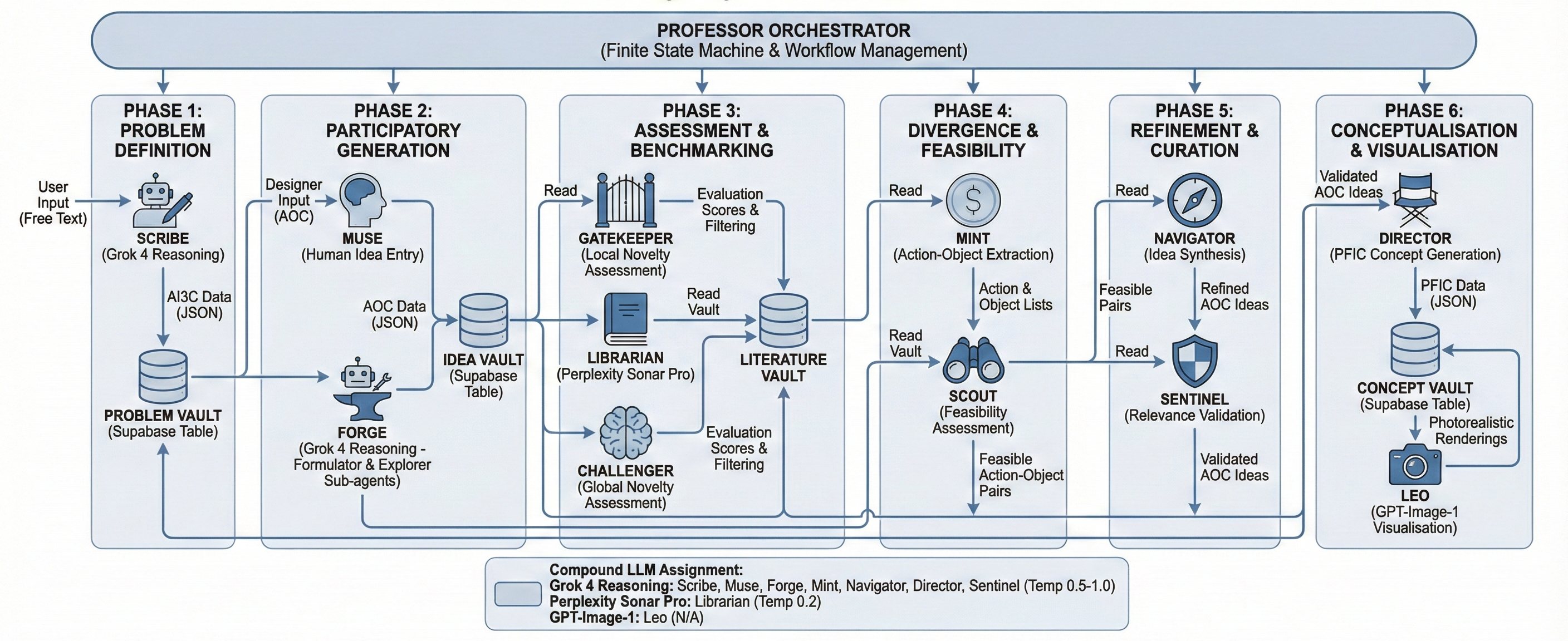}
  \caption{Human-AI Co-creation environment of MIDAS (Meta-Cognitive Ideation using Distributed Agentic AI System) framework. The illustration depicts the collaborative interface where a user interacts with a diverse ensemble of specialized AI agents to facilitate progressive ideation. The distributed agent architecture of MIDAS, illustrating the thirteen specialized agents organized across six operational phases, connected through persistent Vaults and orchestrated by the Professor agent.}
  \Description{This visual teaser illustrates the MIDAS (Meta-Cognitive Ideation using Distributed Agentic AI System) framework, depicting a human creator collaborating with a specialized ensemble of AI agents. The image features a user interfacing with a creative system on the left, juxtaposed against a diverse "team" of distinct AI agents depicted as characters on the right. Each agent is visually unique and equipped with specific tools, symbolizing their distributed cognitive roles in facilitating the progressive ideation process central to the paper.}
  \label{fig:teaser}
\end{teaserfigure}


\maketitle


\section{Introduction}
\label{sec:introduction}


Ideation is the cognitive process of generating, developing, and communicating novel ideas, which is the crucial part of the engineering design process \cite{Beitz1996}. The efficacy of this phase is dependent on the intrinsic expertise, domain knowledge, and tacit experience of human designers. Traditional ideation methods, such as brainstorming \cite{Sutton1996}, synectics \cite{Tang2012}, TRIZ \cite{Terninko1998, Lee2024GeneratingTRIZ}, etc., were developed to provide procedural support; yet, the fundamental creative drive remains as a function of human cognition. This reliance on the individual or group intellect has its own limitations, such as cognitive biases, design fixation \cite{Wadinambiarachchi2024EffectsGenAI}, production blocking in group settings, and the sheer impossibility for any single human to possess a truly encyclopedic knowledge of disparate domains.

\begin{figure*}[ht!]
    \centering
    \includegraphics[width=0.8\linewidth]{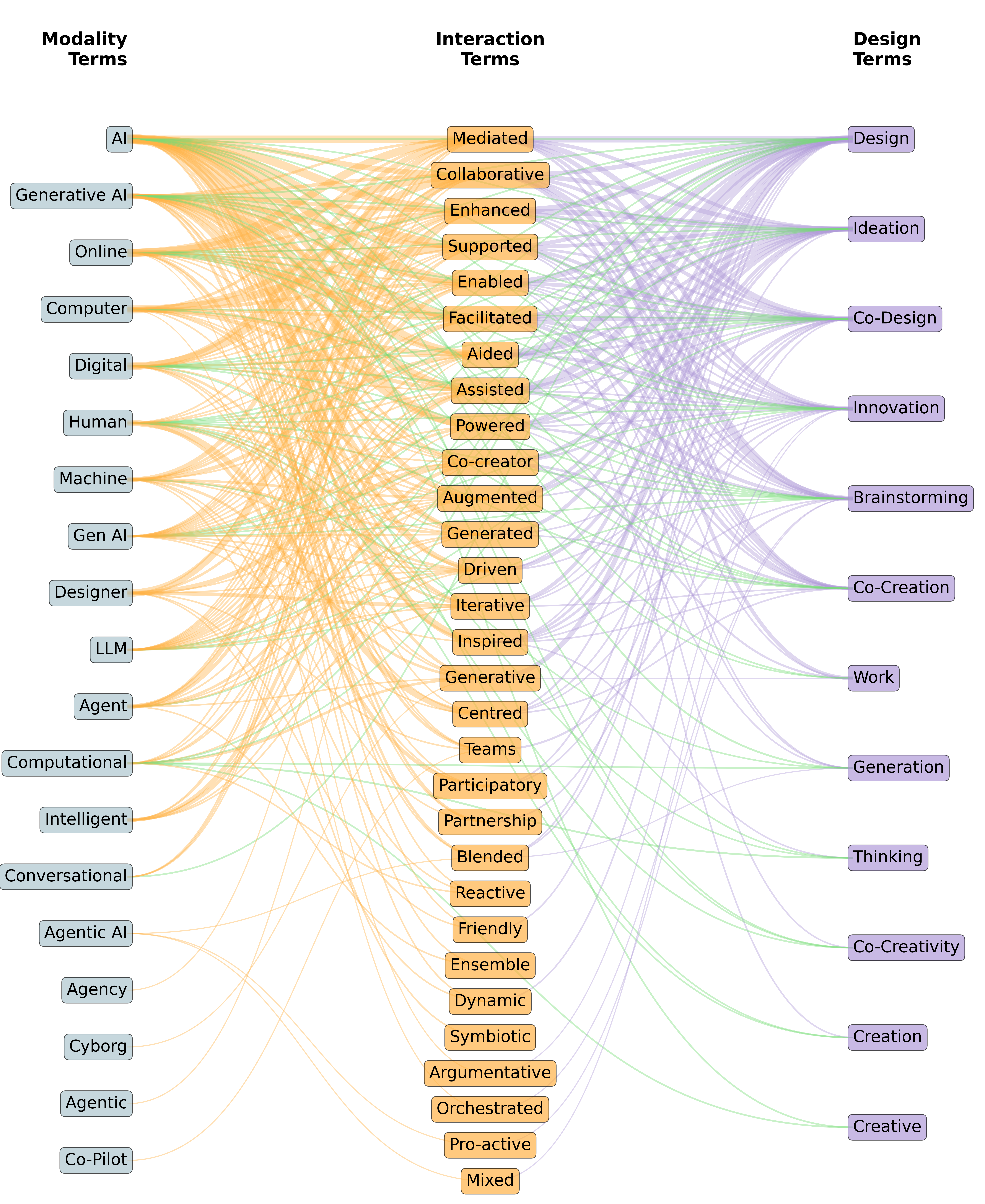}
    \caption{Sankey Map illustrating the interconnections between modality terms (e.g., AI, Generative AI, LLM), interaction terms (e.g., Collaborative, Augmented, Symbiotic), and design terms (e.g., Ideation, Co-Design, Innovation) in AI-assisted creative processes.}
    \label{fig:sankey_map}
\end{figure*}

The need to generate a prolific, novel, and diverse set of ideas swiftly, particularly within the complex, multidisciplinary contexts of modern product design, has necessitated the exploration of computational support \cite{Tsang2022AIIndustrialDesign, Sreenivasan2024DesignThinkingAI}. This quest to augment or assist human creativity with machine intelligence is not new; it has been ongoing for decades under various terminologies (Figure\ref{fig:sankey_map}), such as \textit{Computer-Aided Ideation (CAI)}, \textit{Human-AI Collaborative Ideation}, \textit{Co-ideation}, \textit{AI-Augmented Ideation}, \textit{Co-Creation}, \textit{Human-Machine Collaboration}, \textit{Participatory Design}, \textit{Computational Creativity}, \textit{Mixed-Initiative Ideation}, etc. \cite{Pescher2025RoleAIIdeation, DeFreitas2025IdeationGenAI}. This terminological diversity, as illustrated in Figure \ref{fig:sankey_map}, reflects an academic discourse centered around the relationship between humans and computers \cite{Pescher2025RoleAIIdeation, DeFreitas2025IdeationGenAI}.

Irrespective of the precise term used, the fundamental role of the machine in this partnership is to alleviate the creative burden of the human designer.  Early iterations of CAI tools functioned as passive repositories or simple facilitators, managing morphological analyses or combinatorial explorations. The creative burden remained with the human designer. The exponential rise and democratization of Generative AI, particularly Large Language Models (LLMs) since the 2020s, has shown potential to generate a large number of ideas at an instance given an input problem statement \cite{Wang2025AIProductDesignReview, Abrusci2025AI4Design, Sankar2025a, LLMDesign2025, Zhou2025LLMFunctionsDesign, Wang2025HumanAICoideation, Wang2025CoIdeationFramework}. The machine has transitioned from a passive \textit{tool} to an active \textit{collaborator}, capable of contributing generative output that is often indistinguishable from, and sometimes superior to, that of a human \cite{Zhou2025LLMFunctionsDesign, Choudhury2025AIinDesignProcess, Wang2025ExploringCreativity, MaLLMDiverse2025, Sankar2025a}.

However, the sheer volume of ideas creates a challenge for human experts when assessing their quality, as the process is resource-intensive, cognitively demanding, and difficult to maintain consistency in evaluation criteria. An earlier work has demonstrated that this evaluation is achievable using vector embeddings and clustering \cite{Sankar2025b}. It has been shown that AI-generated ideas tend to form semantic clusters centred around a few foundational concepts, failing to provide the true novelty and diversity that designers require \cite{Sankar2025b}. This paper's primary objective is to address this gap by proposing a system capable of generating a significant number of truly novel ideas. We propose a two-fold definition of novelty: an idea must be distinct from existing solutions available in the literature (global novelty), and it must also be distinct from other ideas generated within the same creative session (local novelty). We argue that all ideation, whether human or computational, flows in a temporal order. Therefore, the $(n+1)^{th}$ idea must be novel not only relative to the literature (existing solutions) but also relative to the $n$ ideas generated before it. This temporal-aware requirement is the principal limitation of existing AI-based ideation systems, which operate on a "single spurt" model, producing a large batch of ideas instantaneously. We propose to exploit this very limitation. In this paper, we propose, design, and implement a new framework based on an agentic AI system, which is explicitly architected to manage this progressive, temporal, and continuous generative-evaluative workflow in ideation.

\section{Potential for Progressive Ideation}
The promise of a true LLM-driven ideation is being hindered by several issues stemming from the single-spurt, monolithic systems. To build a system that genuinely collaborates with a designer, we must first articulate and address these critical shortfalls.



\subsection{The 'Single-Spurt' Limitation in Current AI-Aided Ideation}
\label{sec:intro_single_step}

The contemporary research exploring LLMs in design is burgeoning \cite{Ahmed2025LLMUIUXReview}. The predominant interaction model, however, remains simplistic. The majority of studies and commercial tools frame the human-AI collaboration as a dyadic conversation: a designer engages with a single, monolithic LLM (such as GPT, Grok, or Gemini, etc.) in a turn-by-turn dialogue \cite{Bonnardel2020, Kwon2024DesignerGenerativeAI}. This interaction is often analogized to brainstorming with a partner or a friend who possesses an exceptionally broad knowledge base and the ability to rapidly connect disparate concepts, thereby stimulating the human designer's own creative pathways.

Whilst this conversational paradigm has proven effective for basic idea generation and has been shown to increase the sheer quantity of ideas \cite{Sankar2025a}, it suffers from a fundamental flaw: it relies on a \textit{single-spurt, query-response model}. The interaction is transactional; the designer provides a prompt, and the LLM returns a "spurt" of ideas in an instance. This model is a gross oversimplification and fails to capture the intricate, reflective, and \textit{progressive} nature of genuine human ideation.

Human designers do not generate final, robust concepts in a single flash of insight. The process is inherently iterative, messy, and incremental. It involves a complex interplay of meta-cognitive activities: defining the problem, generating initial thoughts, evaluating them against constraints, researching external stimuli, re-framing the problem based on new insights, and progressively refining a "fuzzy" notion into a concrete, actionable solution in multiple steps. The current single-LLM-as-partner model is incapable of replicating this structured, multi-stage cognitive journey. It operates as a generalist "\textit{answer machine}" rather than a specialized design system, leading to several critical deficiencies that inhibit the generation of truly novel and impactful solutions \cite{Chiarello2024LLMEngineeringDesign, Nguyen2025LLMEngineeringReview}.

\subsection{From Group Dynamics to Monolithic Models}
\label{sec:intro_critique_group}

In professional design practice, ideation is rarely a solo endeavour. It is often conducted in groups to leverage the cognitive diversity of its members. A successful human brainstorming team involves multiple people collaborating with each other, critiquing one another, and actively participating to generate a potential set of viable solutions \cite{Sutton1996}. Each member in the team is trained in unrestrained \textit{generation}, critical \textit{analysis}, and \textit{feasibility}, \textit{researching} prior art, and \textit{synthesizing} disparate ideas into a coherent whole. This dynamic interplay of activities of a collective group of people is what drives innovation \cite{Liu2025PersonaFlow, Obieke2025FrameworkAICED}. The current LLM models discards this rich dynamic, replacing it with a single, generalist "partner". Whilst human groups can suffer from social-cognitive issues like authoritarian bias or production blocking, the solution is not to eliminate the group dynamic but to optimize it. We posit that a far more powerful paradigm exists: a hybrid group comprising a human designer and a team of multiple, specialized AI agents. In such a system, each agent is configured for a specific function. This is the foundation of an \textit{Agentic AI System}, which moves from a simple dyadic partnership to a distributed, multi-agent collaborative structure that mimics a high-performance human design team.

\subsection{Stage-Gate versus Continuous Generation and Assessment}
\label{sec:intro_critique_cgca}
Most AI ideation tools prioritize generation, producing an abundance of ideas \cite{Sankar2025a}, but lack evaluation support. Designers manually filter them for novelty and feasibility, merely shifting the burden, not co-creating. Some machine evaluation works exist, but they are typically used independently after generation. In software engineering, the advent of Continuous Integration/Continuous Deployment (CI/CD) revolutionized development by creating tight feedback loops between writing, testing, and deploying code. We propose that a similar model is necessary for design ideation: a \textit{CG/CA (Continuous Generation / Continuous Assessment)} pipeline. In this model, generation and assessment are not discrete, sequential stages, but are deeply intertwined in a continuous cycle. Ideas are generated, immediately assessed for novelty and diversity, the insights from which are then used to fuel the next round of generation, and so on. This \textit{iterative refinement} is the only method to ensure the progressive enhancement of the idea pool. A prior research has established a robust computational framework for such an assessment, utilizing vector embeddings and clustering to quantify novelty and diversity through metrics like idea sparsity and cluster sparsity \cite{Sankar2025b}, providing a methodological foundation for the "CA" part of the pipeline.

\subsection{The 'Novelty Illusion': Hallucination, Variants, and Fixation}
\label{sec:intro_critique_novelty}

A significant, and often unreported, issue with current LLMs is the quality of the "novelty" they produce. Many research work claim that LLMs generate more novel ideas than humans, but this claim is often based on skewed samples. LLMs are, by their very nature, stochastic parrots trained on the median of existing human knowledge. They excel at \textit{interpolation}, i.e., creating plausible-sounding variants of existing solutions, but struggle with true \textit{extrapolation}, or radical, out-of-box innovation. This results in a "sea of sameness," where thousands of generated "ideas" are merely minor syntactic or semantic variations of each other, a phenomenon we term the \textit{"novelty illusion."}

Compounding this issue is the lack of real-world grounding. Another flaw in existing AI-based ideation systems is that they generate ideas, with no mechanism to compare them against the vast corpus of existing solutions, such as patents, academic papers, and commercial products. An idea may seem novel to the designer and the LLM, but it may have been patented a decade prior. Our goal, therefore, is not "idea abundance" but to provide the designer with a handful of ideas that are truly novel and diverse, both in relation to each other (local novelty) and in relation to existing solutions (global novelty).

\subsection{Linear Ideation: The Untapped Potential of 'Ideas from Ideas'}
\label{sec:intro_critique_linear}

The prevailing model of AI ideation is linear and unidirectional: a designer inputs a Problem (P1), and the AI generates a list of Ideas \{I1, I2, I3...I5\}. This assumes that ideas are tethered to the problem from which they were spawned. We argue that this approach fails to capture a core mechanism of human creativity: disparate connections, conceptual blending, and analogical transfer. Ideas are not intrinsically bound to a single problem. An idea (say, I4) generated for Problem P1 might also be a viable solution for an entirely different Problem (P2). This new Problem P2, in turn, will have its own constellation of existing solutions \{I4, I5, I6, I7, I8\}. The truly profound creative leap is to then ask: what are the chances that one of these other solutions for P2 (e.g., I6 or I7) could be adapted to solve the original Problem P1? We believe this non-linear, networked approach, as depicted in Figure \ ref {fig:nonlinear-ideation-network} of generating "\textit{ideas from ideas}" allows the designer to break free from the conceptual boundaries of their initial problem, a capability entirely absent from current linear generation models.

\begin{figure*}[ht!]
    \centering
    \includegraphics[width=0.75\textwidth]{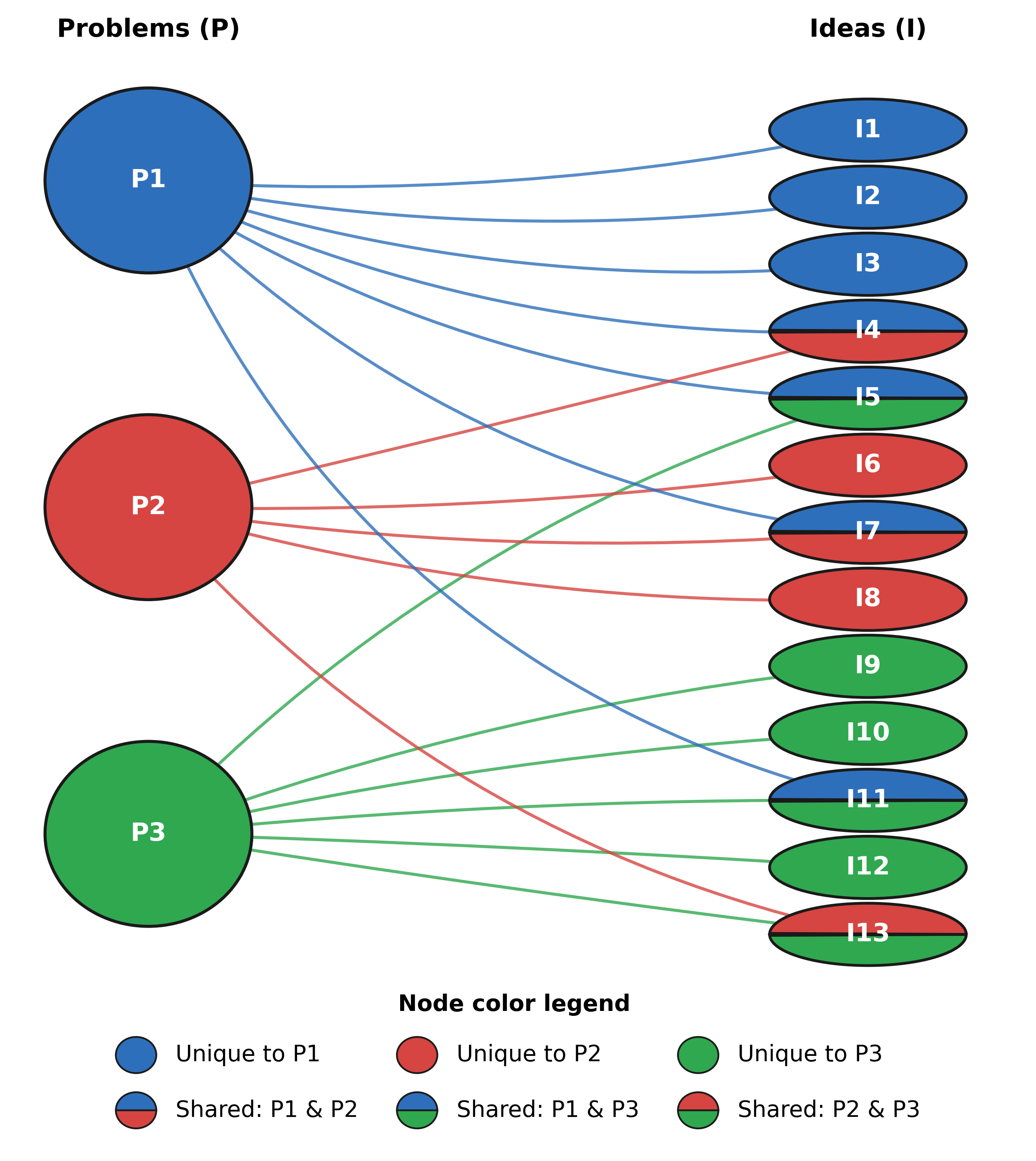}
    \caption{\textbf{Networked, non-linear ideation via “ideas-from-ideas.”}
    Unlike the prevailing linear model, where a single problem $P_1$ yields a one-way list of ideas $\{I_1,\dots\}$, this diagram depicts ideation as a reusable, cross-problem network.
    Problems ($P$) are shown on the left and idea nodes ($I$) on the right, with edges indicating applicability.
    Crucially, some ideas generated for one problem are also viable for other problems (e.g., $I_4$ links $P_1$ and $P_2$; $I_5$ links $P_1$ and $P_3$), enabling conceptual blending and analogical transfer.
    This “re-association” creates pathways where ideas discovered under $P_2$ (e.g., $I_6$--$I_8$) can be adapted back to address $P_1$ (e.g., $I_7$), and similarly across other problem–idea constellations (e.g., $I_{13}$ links $P_2$ and $P_3$, while $I_{11}$ links $P_1$ and $P_3$).
    Node colors encode whether an idea is unique to a single problem or shared across problems, highlighting bridges that enable non-linear creative leaps.}
    \label{fig:nonlinear-ideation-network}
\end{figure*}

\subsection{The 'Evaluation Overload' and Passive Designer Role}
\label{sec:intro_passive_role}
Designers often do not prefer AI in their creative endeavours; this is not Luddism; it is a reaction to a paradigm that actively devalues their contribution. This is because these systems offer no mechanism for integrating the designer's own ideas. The human's creative spark is sidelined, fostering a sense of disengagement and a loss of ownership. This is a psychological barrier to adoption. To overcome this, we propose a \textit{'PAC' (Participatory, Active, Collaborative) Ideation Framework} as detailed in Section:\ref{sec:midas_pac}. In a PAC-based system, the designer is central, not peripheral. They are an active participant who contributes their own ideas to the system as well, rather than just being a prompter. These human-generated ideas are then fed into the same evaluation and development pipeline as the machine-generated ideas. The assessment occurs in a blind review, where the assessors are unaware of or indifferent to whether an idea originated from a human or an AI; all ideas are evaluated on their own merit. This would encourage the designer to remain engaged and create a truly synergistic partnership.

\subsection{The Rationale for Progressive Ideation: An Experimental Insight}
\label{sec:rationale_progressive}

To validate the core hypothesis of this paper, we first conducted a brief experiment that builds upon previously published work. This earlier paper \cite{Sankar2025a} introduced DesignerAI, a fine-tuned LLM that generates an "abundance" of ideas in a single-spurt. This capability was tested across six distinct problem statements (PS1 to PS6) covering a wide range of domains.

\begin{itemize}
\item \textbf{PS1:} Product for segregation as a means for effective waste management.
\item \textbf{PS2:} Product for footwear disinfection and cleaning for improved hygiene and safety.
\item \textbf{PS3:} Product for enhancing household dish cleaning efficiency and sustainability.
\item \textbf{PS4:} Product for enhancing comfort and efficiency for prolonged standing in queues.
\item \textbf{PS5:} Product for bird-feeding for fostering mental well-being of elderly individuals at home.
\item \textbf{PS6:} Product for convenient umbrella drying and storage on travel.
\end{itemize}

\begin{figure*}[ht!]
    \centering
    \begin{subfigure}[b]{0.44\textwidth}
        \includegraphics[width=\linewidth]{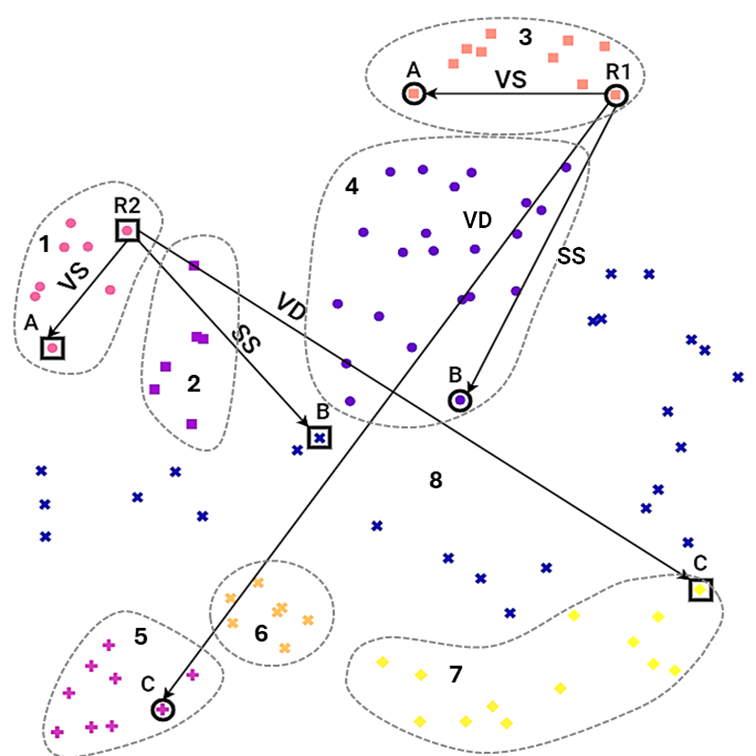}
        \caption{PS1: Single-Spurt Clusters}
        \label{fig:ps1_ss}
    \end{subfigure}
    \hfill
    \begin{subfigure}[b]{0.44\textwidth}
        \includegraphics[width=\linewidth]{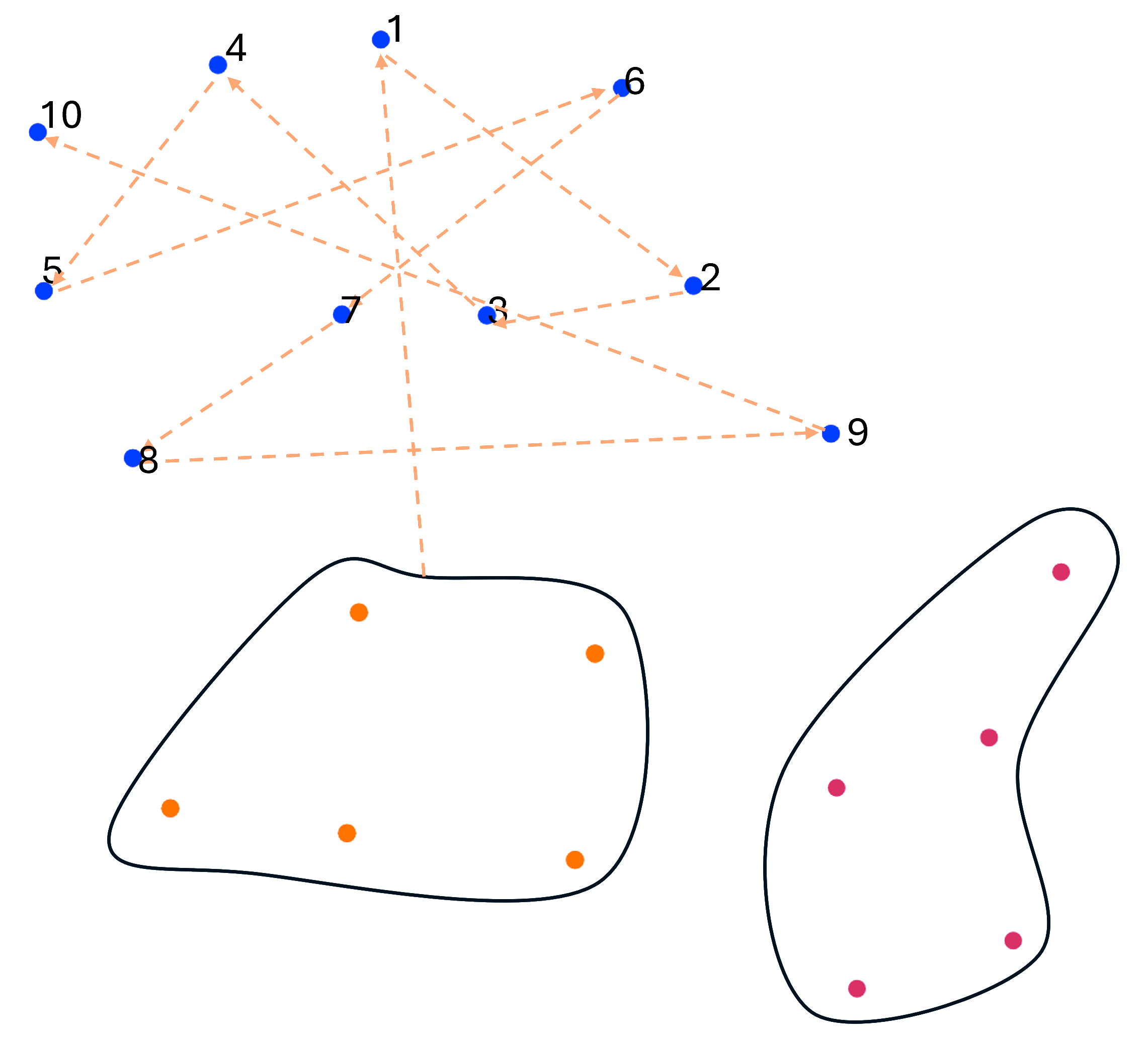}
        \caption{PS1: Progressive Provocation}
        \label{fig:ps1_pp}
    \end{subfigure}

    \vspace{0.5cm}

    \begin{subfigure}[b]{0.44\textwidth}
        \includegraphics[width=\linewidth]{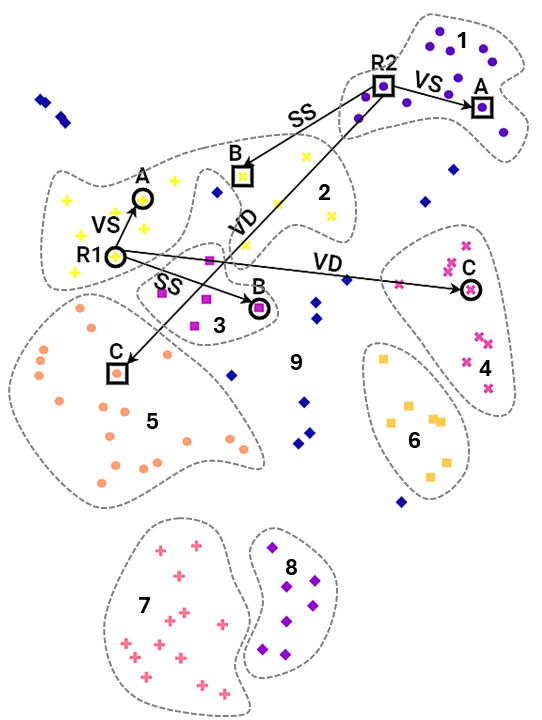}
        \caption{PS2: Single-Spurt Clusters}
        \label{fig:ps2_ss}
    \end{subfigure}
    \hfill
    \begin{subfigure}[b]{0.48\textwidth}
        \includegraphics[width=\linewidth]{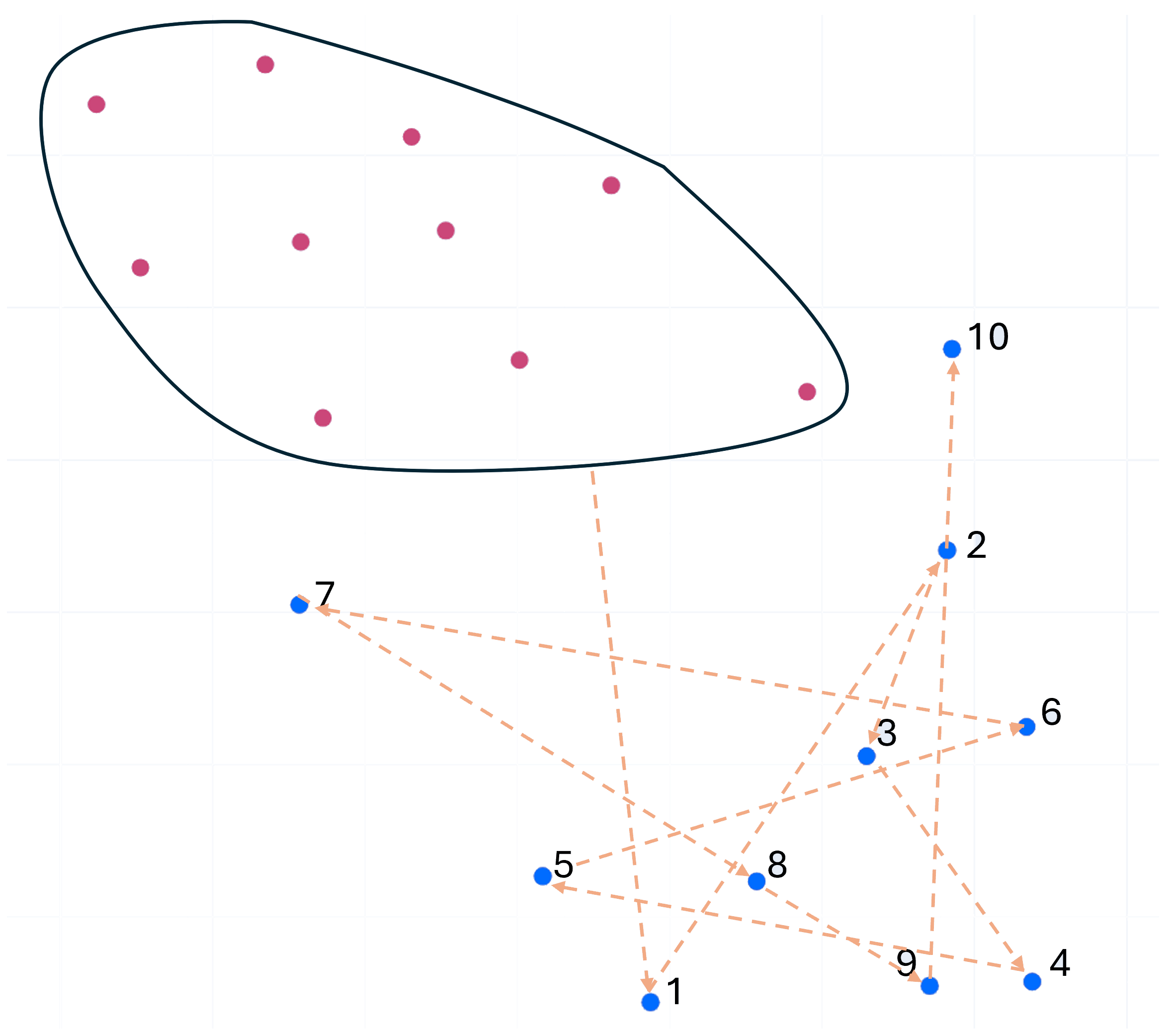}
        \caption{PS2: Progressive Provocation}
        \label{fig:ps2_pp}
    \end{subfigure}

    \caption{A comparative visualization of semantic clustering for idea generation methodologies for the problem statement PS1 (Waste Segregation) and PS2 (Footwear Disinfection). (a), (c) 'Single-spurt' generation results in high-density clusters of similar ideas. (b), (d) 'Progressive provocation' forces the generation of semantically distant and diverse ideas (blue dots). The transition from high-density clusters in the baseline to diverse micro-clusters in the progressive sets is evident across both domains.}
    \label{fig:results_ps1_ps2}
\end{figure*}

\begin{figure*}[ht!]
    \centering
    \begin{subfigure}[b]{0.35\textwidth}
        \includegraphics[width=\linewidth]{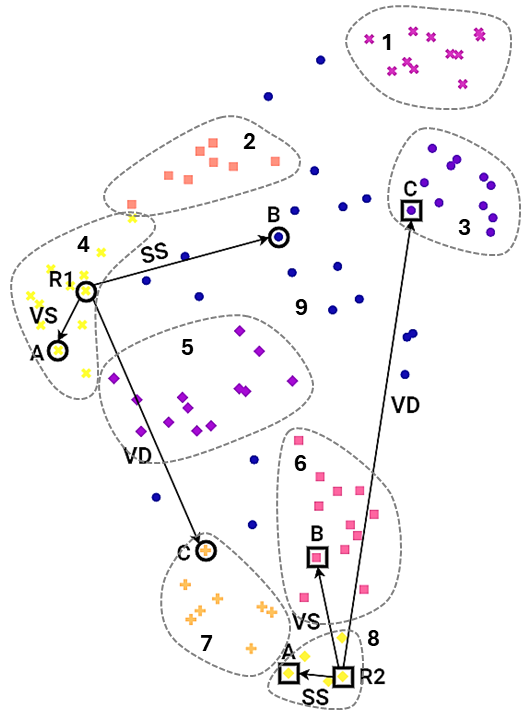}
        \caption{PS3: Single-Spurt Clusters}
        \label{fig:ps3_ss}
    \end{subfigure}
    \hfill
    \begin{subfigure}[b]{0.44\textwidth}
        \includegraphics[width=\linewidth]{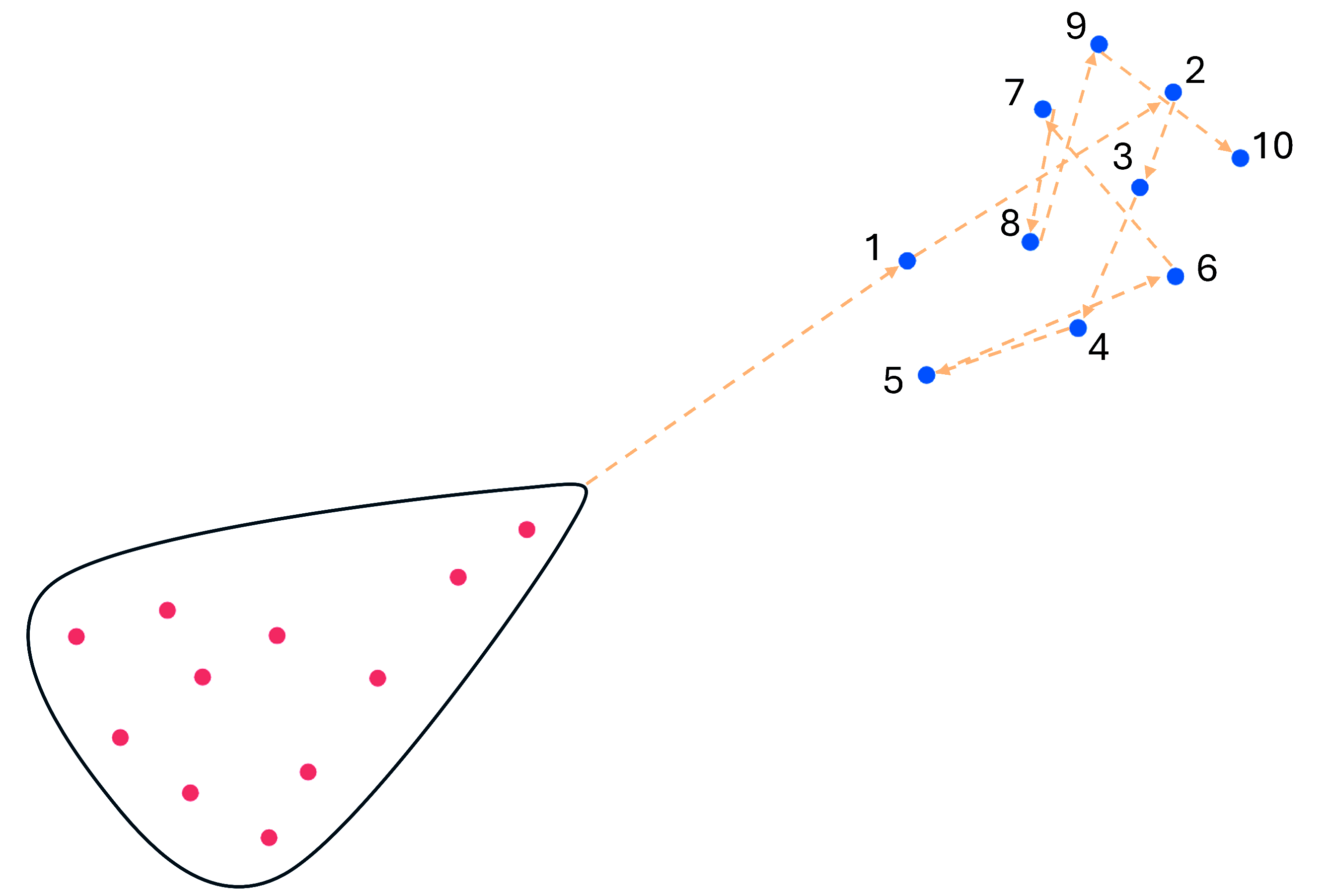}
        \caption{PS3: Progressive Provocation}
        \label{fig:ps3_pp}
    \end{subfigure}

    \vspace{0.5cm}

    \begin{subfigure}[b]{0.4\textwidth}
        \includegraphics[width=\linewidth]{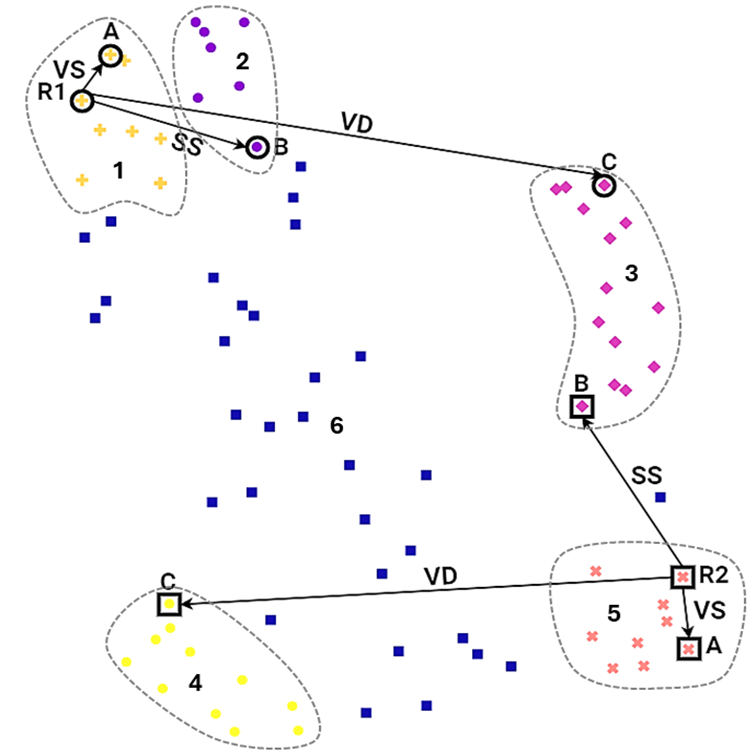}
        \caption{PS4: Single-Spurt Clusters}
        \label{fig:ps4_ss}
    \end{subfigure}
    \hfill
    \begin{subfigure}[b]{0.35\textwidth}
        \includegraphics[width=\linewidth]{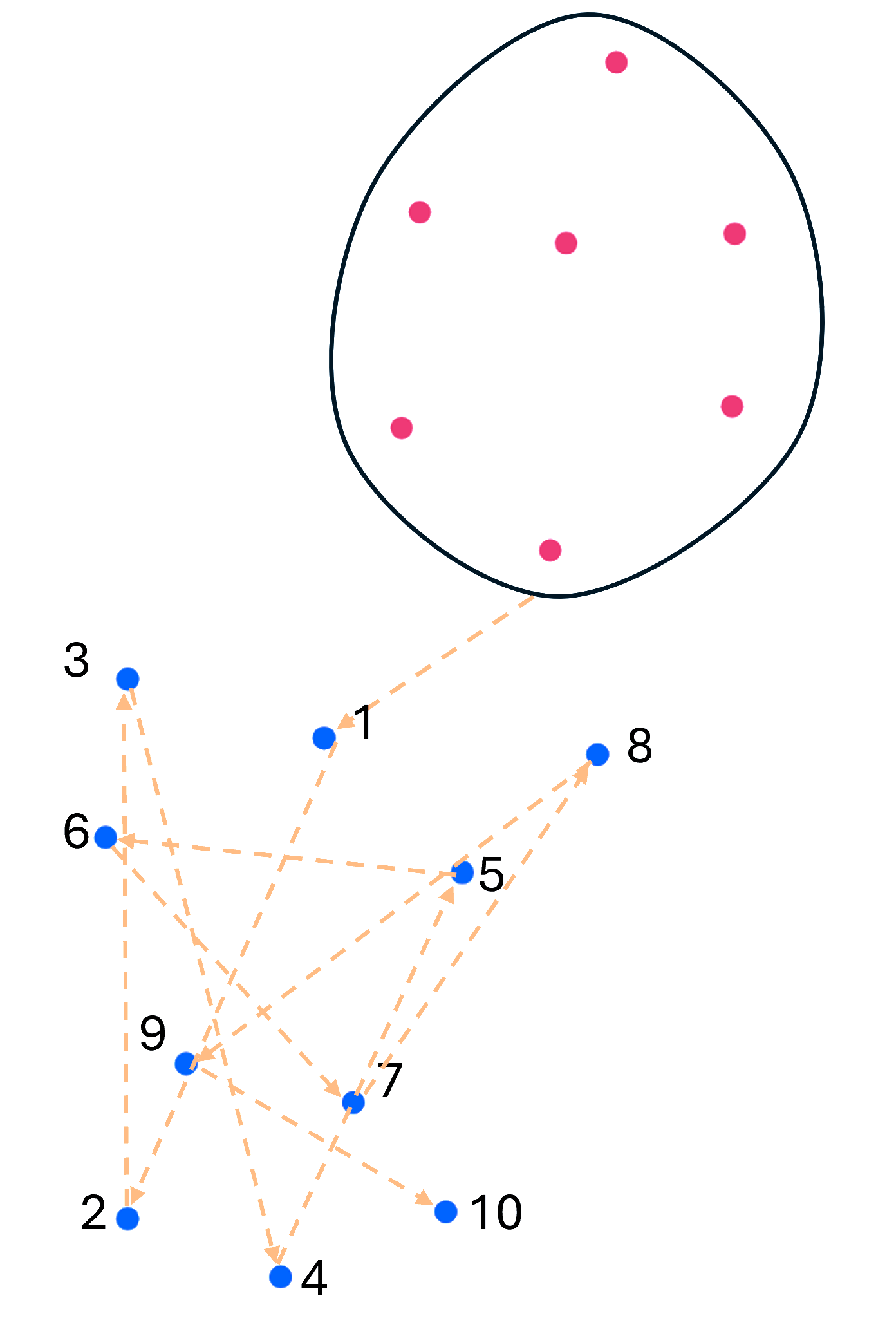}
        \caption{PS4: Progressive Provocation}
        \label{fig:ps4_pp}
    \end{subfigure}

    \caption{A comparative visualization of semantic clustering for idea generation methodologies for the problem statement PS3 (Dish Cleaning) and PS4 (Queuing Comfort). (a), (c) 'Single-spurt' generation results in high-density clusters of similar ideas. (b), (d) 'Progressive provocation' forces the generation of semantically distant and diverse ideas (blue dots). The transition from high-density clusters in the baseline to diverse micro-clusters in the progressive sets is evident across both domains.}
    \label{fig:results_ps3_ps4}
\end{figure*}

\begin{figure*}[ht!]
    \centering
    \begin{subfigure}[b]{0.44\textwidth}
        \includegraphics[width=\linewidth]{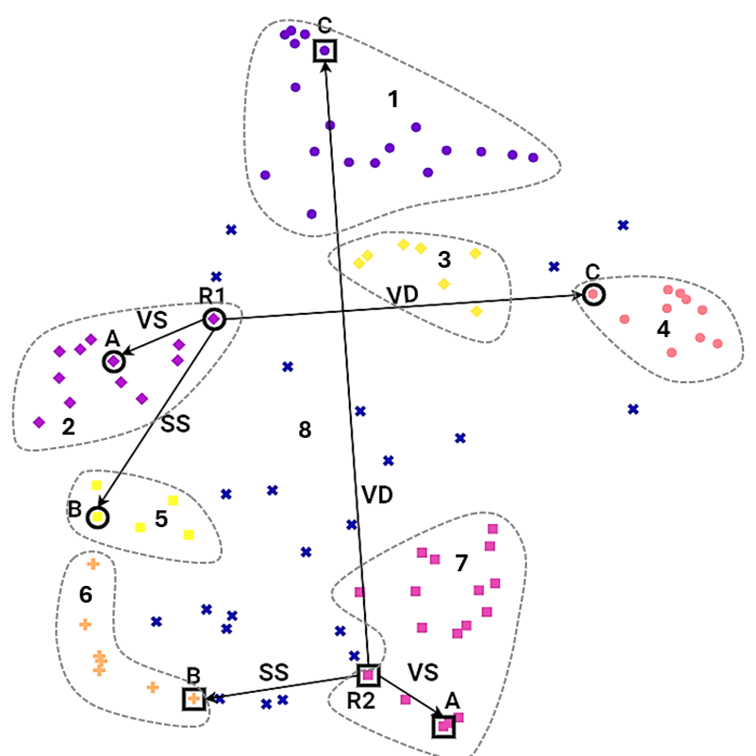}
        \caption{PS5: Single-Spurt Clusters}
        \label{fig:ps5_ss}
    \end{subfigure}
    \hfill
    \begin{subfigure}[b]{0.35\textwidth}
        \includegraphics[width=\linewidth]{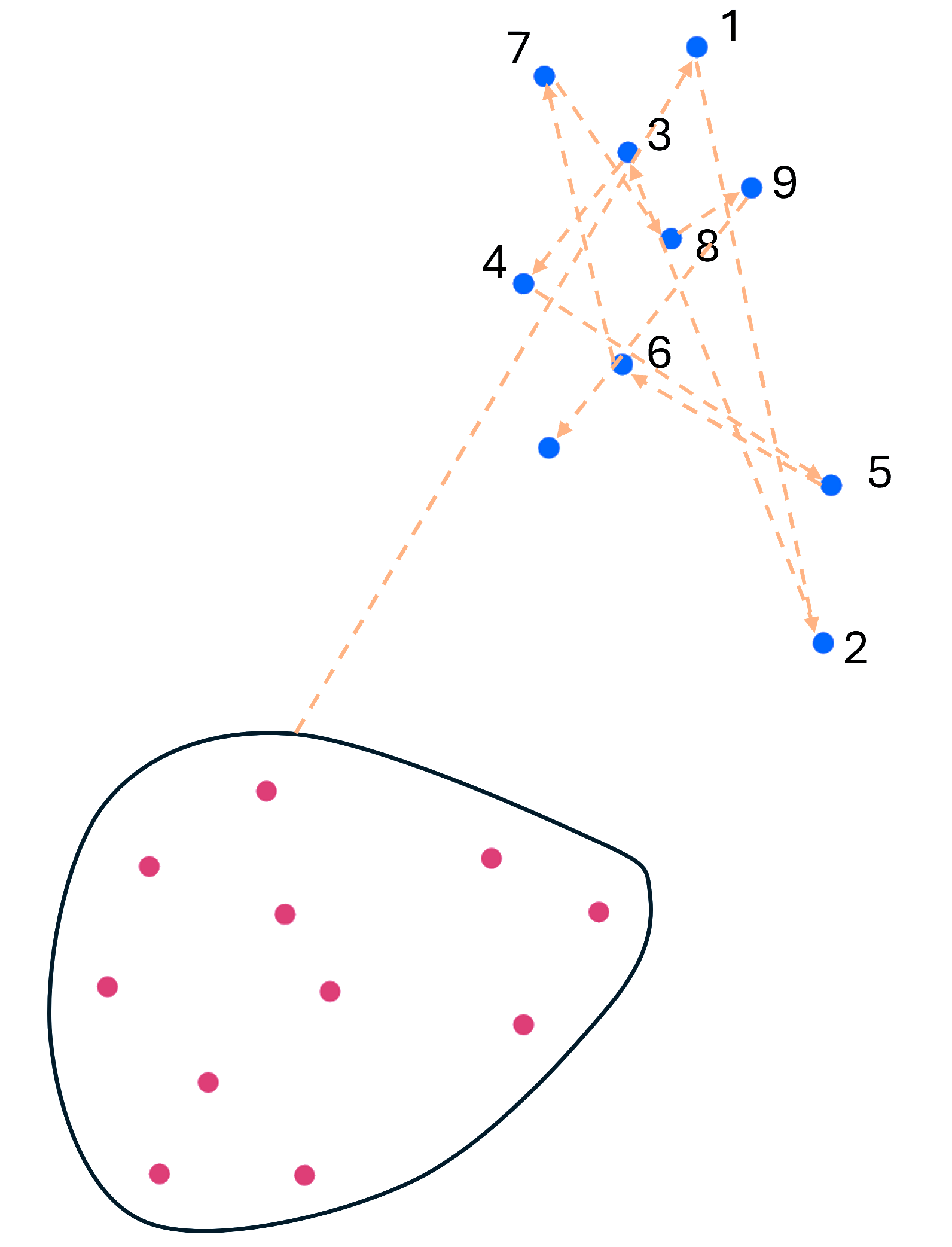}
        \caption{PS5: Progressive Provocation}
        \label{fig:ps5_pp}
    \end{subfigure}

    \vspace{0.5cm}

    \begin{subfigure}[b]{0.4\textwidth}
        \includegraphics[width=\linewidth]{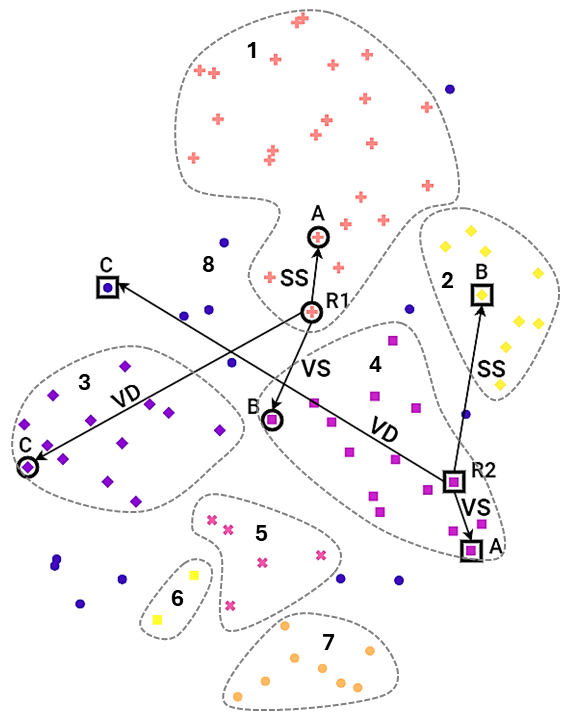}
        \caption{PS6: Single-Spurt Clusters}
        \label{fig:ps6_ss}
    \end{subfigure}
    \hfill
    \begin{subfigure}[b]{0.4\textwidth}
        \includegraphics[width=\linewidth]{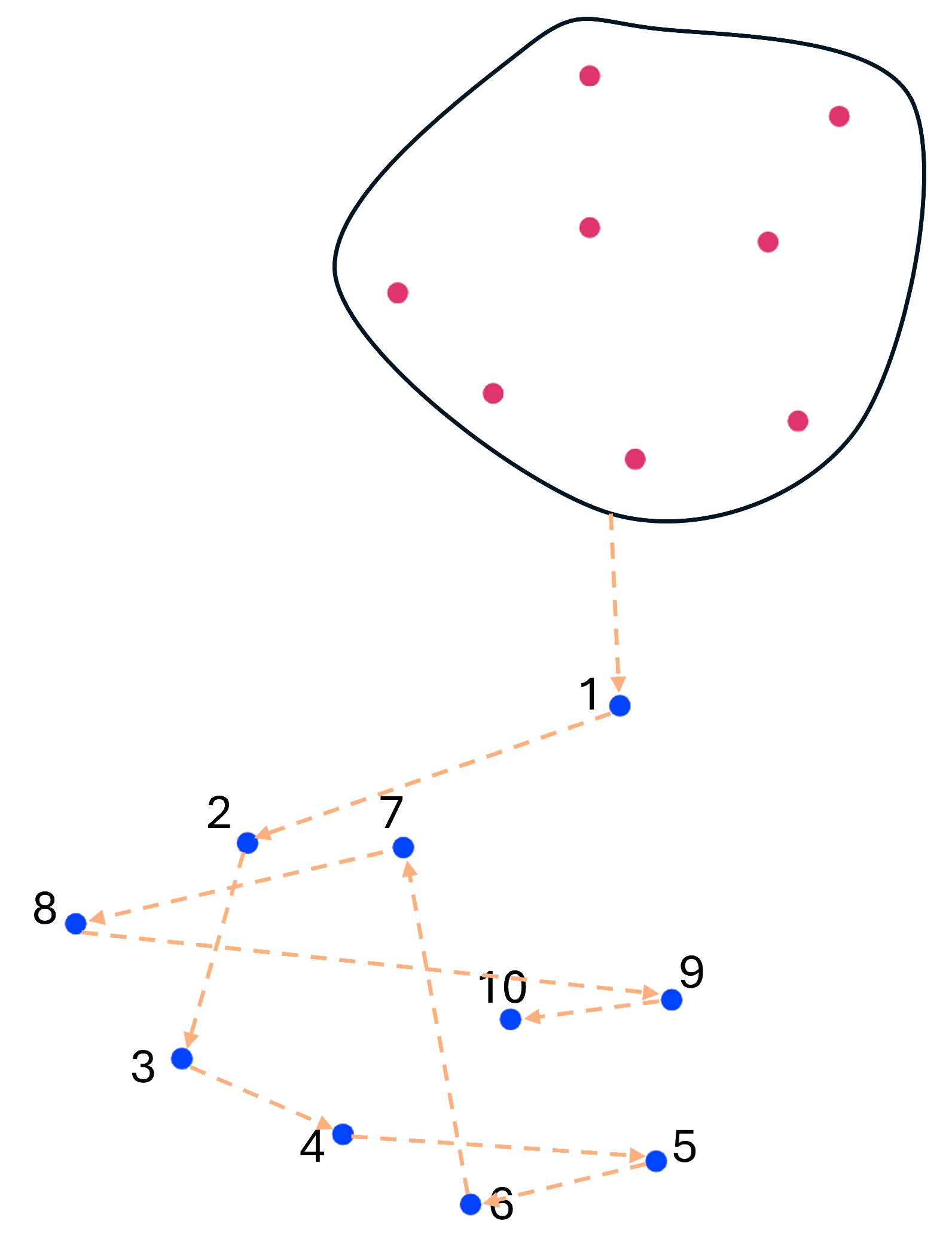}
        \caption{PS6: Progressive Provocation}
        \label{fig:ps6_pp}
    \end{subfigure}

    \caption{A comparative visualization of semantic clustering for idea generation methodologies for the problem statement PS5 (Bird Feeding) and PS6 (Umbrella Storage). (a), (c) 'Single-spurt' generation results in high-density clusters of similar ideas. (b), (d) 'Progressive provocation' forces the generation of semantically distant and diverse ideas (blue dots). The transition from high-density clusters in the baseline to diverse micro-clusters in the progressive sets is evident across both domains.}
    \label{fig:results_ps5_ps6}
\end{figure*}

For each problem statement, DesignerAI generated 100 ideas. In a subsequent work \cite{Sankar2025b}, these idea sets were computationally assessed using a novel framework based on vector embeddings and clustering. The cluster plots for a representative sample of these domains are reproduced in the left column of Figures \ref{fig:results_ps1_ps2}, \ref{fig:results_ps3_ps4} and \ref{fig:results_ps5_ps6}. As the visualizations show, across all six domains, the 100 ideas converged into distinct, high-density clusters. Ideas within each cluster are semantically Very Similar (VS), while the clusters themselves are Very Different (VD). This empirically demonstrates a critical limitation of single-spurt generation: LLM-based ideation produces a large volume of ideas, but some of those ideas themselves are semantic variants, closely centred around a few core novel ideas.

We then sought to determine if an LLM could be 'provoked' to overcome this limitation and generate ideas progressively. For each of the six problem statements, we first prompted the DesignerAI to generate 10 ideas in a single shot. Following this, the model was iteratively prompted to generate one new idea at a time, for 10 subsequent iterations. A critical constraint was applied at each step: the new idea must be semantically distinct from all previously generated ideas (the initial 10 plus all progressive ideas), which were stored in its contextual buffer memory for recall.

These 20 total ideas per PS (10 single-spurt + 10 progressive) were then processed using the same clustering algorithm from \cite{Sankar2025b}, with the results visualized in the right column of Figures \ref{fig:results_ps1_ps2}, \ref{fig:results_ps3_ps4} and \ref{fig:results_ps5_ps6}. Across all cases, the plots clearly show the initial ideas (represented by the grouped pink or orange dots) forming dense clusters, confirming our hypothesis about single-spurt generation. In stark contrast, the 10 progressively generated ideas (blue dots, numbered 1 to 10) are consistently thrown away from the initial clusters at a significant semantic distance, each forming its own distinct micro-cluster.

While these results demonstrate that LLMs have the latent potential to generate truly novel and diverse ideas when nudged through iterative provocation, this prompt-based nudging is not without its limitations. Relying solely on manual prompting is labor-intensive and lacks structural grounding; it relies on the model’s internal "recall" without a formal mechanism to benchmark against global prior art or technical feasibility. Furthermore, as the number of iterations increases, simple prompts often fail to maintain semantic distance, leading to "hallucinated novelty" or repetitive variants. This experiment provided the foundational motivation for our work, showing that while progressive ideation is possible, it necessitates a formal, multi-agent framework to systematically manage the generation, evaluation, and refinement of ideas in a scalable and professional design context. Thus, this paper seeks to bridge this gap by introducing the concept of \textit{Progressive Ideation} through an agentic AI framework.

Progressive Ideation is an incremental and iterative journey from a fuzzy, ill-defined problem-space to a set of concrete, well-defined solutions. The progressive ideation operationalizes the 'CG/CA' pipeline such that each generation is novel compared to the last generation and solves the 'abundance problem' by delivering a curated set of diverse ideas rather than a thousand variants. Therefore, we believe the prevailing paradigm of human-AI ideation, built on single-spurt, dyadic, and monolithic LLM interactions, is flawed. It fails to capture the multi-stage, evaluative, and incremental nature of human creativity, relegating the human designer to a passive, peripheral role. 

Our primary objectives are:
\begin{enumerate}
    \item To design and develop a novel agentic AI architecture that decomposes the complex act of ideation into a distributed system of specialized, collaborative agents, moving beyond the single-LLM paradigm.
    \item To implement a 'PAC' (Participatory, Active, Collaborative) framework within this architecture, centering the human designer as an active co-creator whose contributions are integrated and valued.
    \item To operationalize a 'CG/CA' (Continuous Generation / Continuous Assessment) pipeline that facilitates true progression, moving from fuzzy problems to concrete concepts incrementally.
    \item To integrate a global and local novelty check within this pipeline, benchmarking generated ideas against real-world patents and products and against themselves to filter for true innovation over simple variants.
\end{enumerate}


\section{A Framework for Progressive Human-AI Ideation: MIDAS}
\label{sec:midas_framework}
\begin{sloppypar}
To address these objectives, we present a novel system named \textit{Meta-cognitive Ideation through Distributed Agentic AI System (MIDAS)}\footnote{The system is accessible online via \url{https://midas-ai-ideation.vercel.app/}. Access credentials to try out MIDAS will be provided by sending a request to us}. MIDAS is a distributed, multi-agent system comprising 13 distinct, specialized agents. These agents work in concert with a human designer to manage the entire ideation workflow, from structuring an ill-defined problem to rendering photorealistic concepts of the final, curated ideas. To manage the complex flow of information between these agents, the framework is built upon a prior foundational work in structured knowledge representation for design communication \cite{Sankar2025a}. This work established a novel typology for design artefacts: the \textit{AI3C (Activity, Item, Contradiction, Criteria, Constraints)} model for representing problem statements, the \textit{AOC (Action, Object, Context)} model for representing ideas, and the \textit{PFIC (Principle, Features, Implementation, Characteristics)} model for representing concepts. This shared, structured "language" allows the specialized agents to collaborate seamlessly, passing information with high fidelity and enabling the complex, multi-stage reasoning.
\end{sloppypar}


\begin{figure*}[t!]
    \centering
    \includegraphics[width=0.7\linewidth]{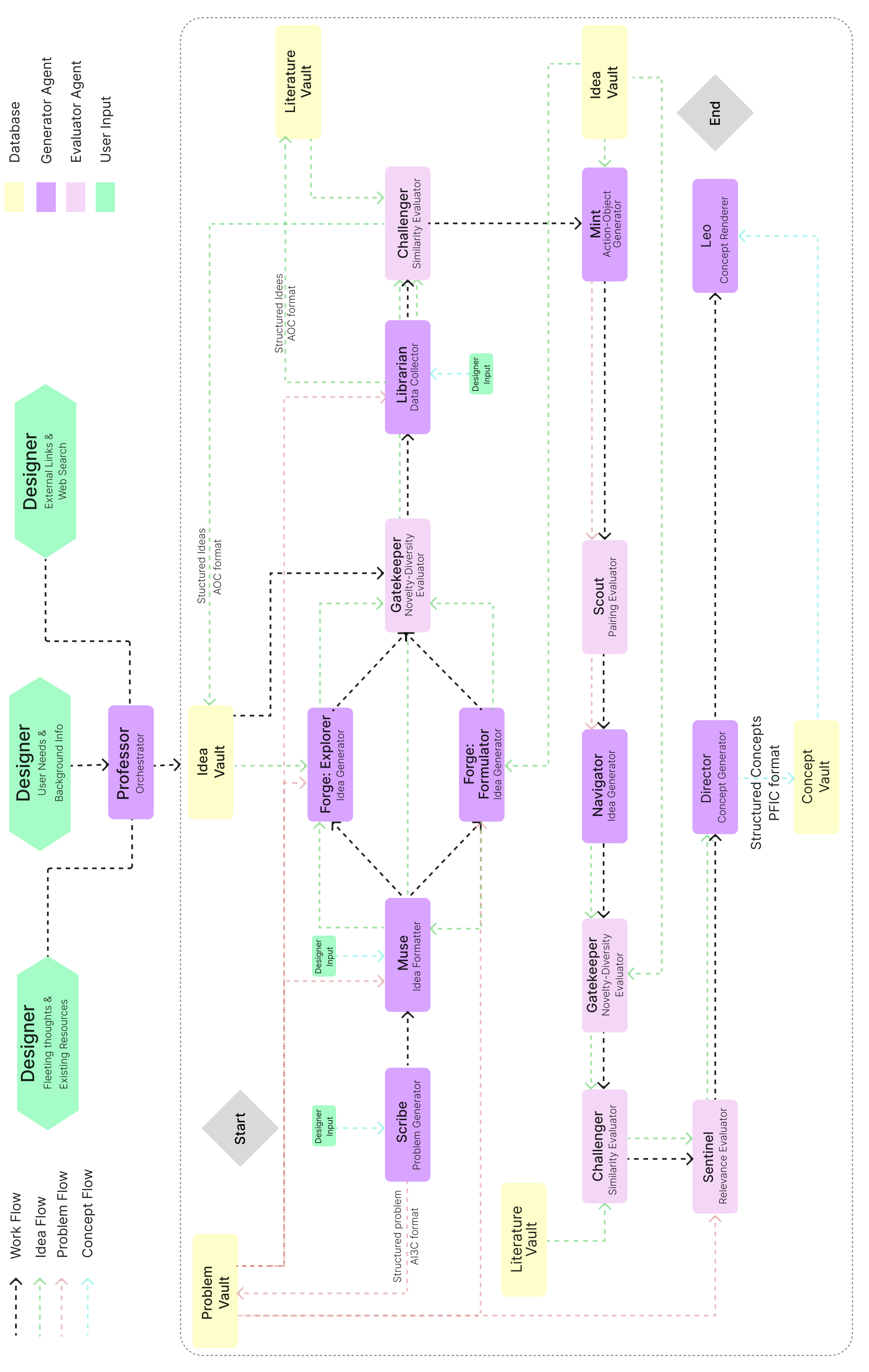} 
    \caption{The Detailed System Architecture of the MIDAS Framework, illustrating the flow of work, ideas, and concepts between Generator Agents, Evaluator Agents, persistent Vaults, and the human Designer.}
    \label{fig:midas_architecture}
\end{figure*}

\subsection{Emulating the Human Cognitive Process through Agents}
\label{sec:midas_cognitive_emulation}

Human ideation, particularly in complex engineering design, is not a singular event but a sophisticated cognitive process involving a sequence of distinct meta-cognitive activities, which we capture through different agents as follows:

\begin{itemize}
    \item \textbf{Scribe} -- \textit{Define:} Understanding user needs and synthesizing insights to frame an actionable, structured problem statement.
    
    \item \textbf{Muse \& Forge} -- \textit{Brainstorm \& Ideate:} Generating a broad range of initial ideas and systematically generating solutions from the given problem statement.
    
    \item \textbf{Gatekeeper \& Scout} -- \textit{Evaluate:} Assessing the generated ideas against constraints: novelty, and feasibility.
    
    \item \textbf{Librarian \& Challenger} -- \textit{Gather \& Compare:} Comparing generated solutions against existing knowledge or prior art.
    
    \item \textbf{Mint} -- \textit{Diverge:} Pushing the boundaries of the solution space to discover semantically novel pathways from existing ideas.
    
   \item \textbf{Navigator \& Sentinel} -- \textit{Refine \& Curate:} Adjusting, blending, and polishing ideas through analogical reasoning and structured refinement.
    
    \item \textbf{Director \& Leo} -- \textit{Create \& Conceptualize:} Synthesizing final ideas into a coherent, detailed, and actionable concept proposal and representational images.
\end{itemize}

The architecture of MIDAS, visualized in Figure \ref{fig:midas_architecture}, consists of a distributed system of agents, as listed above, connected by a series of persistent databases known as "Vaults." The workflow is managed by an orchestrator (termed '\textbf{Professor}' agent) that directs the flow of data between agents. This architecture explicitly implements the \textbf{ReACT (Reasoning + Acting)} paradigm \cite{yao2023reactsynergizingreasoningacting, zhou2024languageagenttreesearch, acikgoz2025speakrlsynergizingreasoningspeaking}, where each agent's action is a discrete, reasoned step in a much larger cognitive process.

\subsection{A 'PAC' (Participatory, Active, Collaborative) Ideation Partnership}
\label{sec:midas_pac}

A core philosophical principle of the MIDAS architecture is the establishment of a \textbf{'PAC' (Participatory, Active, Collaborative)} partnership, designed to overcome the "passive curator" role that designers reject in traditional AI tools. MIDAS achieves this primarily through the 'Muse' agent.

The ideation pipeline does not begin with the AI generating solutions. After the user needs and issues are provided by the designer to 'Scribe', which generates structured problem statements, the first generative step is reserved for the human. The 'Muse' agent offers a dedicated interface for designers to input their own initial ideas. These human-generated ideas are then translated into the same structured AOC (Action, Object, Context) format that the AI agent ('Forge') uses.

These human-generated ideas are deposited into the 'Idea Vault' alongside those generated by the AI. The subsequent evaluation agent, 'Gatekeeper', is "blind" to the source. It assesses all ideas—human and AI—using the same impartial computational metrics for novelty and diversity. The designer is free to intervene at any stage and provide their thoughts, rerunning the entire agentic pipeline. The designer is also provided with the authority to add additional ideas that the AI may have discarded and/or remove any ideas that the AI had shortlisted. We believe this mechanism has a profound psychological effect:
\begin{enumerate}
    \item \textbf{Participatory:} The designer is a direct contributor and driver to the idea pool, not just a prompter.
    \item \textbf{Active:} The designer's ideas are not sidelined; they actively compete with and evolve alongside the AI's ideas.
    \item \textbf{Collaborative:} This creates a sense of shared ownership and true co-creation. The final, curated concepts are a hybrid blend of the best ideas, regardless of their origin.
\end{enumerate}
This 'PAC' model ensures the designer remains engaged, central to the process, and respects their creative agency, thereby fostering a much healthier and more productive human-AI partnership.


\subsection{Agent Ensemble: Roles and Responsibilities}
\label{sec:midas_agents}

MIDAS decomposes the ideation process into specialized agents. Each agent is specifically prompted by an LLM designed for a single, discrete task. This specialization, akin to a human design team, allows for a far more robust and sophisticated workflow.

\subsubsection{Phase 1: Problem Definition}
\label{sec:agent_scribe}
\textbf{1. Scribe (Problem Statement Structurer)}
\begin{itemize}
    \item \textbf{Purpose:} To translate the designer's often "fuzzy" and qualitative problem statement into a structured, computationally legible format. A clearly defined problem is the foundation of successful design \cite{Beitz1996}.
    \item \textbf{Input:} Unstructured textual problem statement from the user.
    \item \textbf{Process:} Scribe parses the text to identify and extract key design parameters.
    \item \textbf{Output:} The problem is structured into our novel {AI3C (Activity, Item, Contradiction, Criteria, Constraints)} format \cite{Sankar2025a}. This structured data is saved to the `Problem Vault` for all other agents to access.
    \item \textbf{Example:}
    \begin{quote}
        \textbf{Input:} "Elderly people find it increasingly difficult to sit and stand from a chair... They depend on others..."
        
        \textbf{Output (AI3C Format):}
        \begin{itemize}
            \item \textit{Activity:} Assisting with the transition between sitting and standing.
            \item \textit{Item:} Elderly individuals using chairs.
            \item \textit{Constraints:} Must be safe, easy to use, affordable, compatible with home environments.
            \item \textit{Contradiction:} Users need support to sit/stand independently, but conventional chairs offer no assistance.
            \item \textit{Criteria:} Enables independence, reduces external assistance, maintains comfort.
        \end{itemize}
    \end{quote}
\end{itemize}

\subsubsection{Phase 2: Participatory Generation}
\label{sec:agent_muse_forge}
\textbf{2. Muse (Idea Structurer)}
\begin{itemize}
    \item \textbf{Purpose:} To serve as the primary interface for the human designer's creative contribution, implementing the 'PAC' model.
    \item \textbf{Input:} An unstructured textual description of ideas from the designer.
    \item \textbf{Process:} Muse takes the designer's idea and, referencing the AI3C problem context from `Scribe`, re-frames it into the standardized idea format.
    \item \textbf{Output:} The human idea is structured into the {AOC (Action, Object, Context)} format \cite{Sankar2025a} and deposited in the `Idea Vault`.
    \item \textbf{Example:}
    \begin{quote}
        \textbf{Input:} "A bio-mimetic structure attached to the human body to reduce the muscular load on the knee..."
        
        \textbf{Output (AOC Format):}
        \begin{itemize}
            \item \textit{Idea:} Bio-mimetic Support Exoskeleton
            \item \textit{Action:} Attach and Support
            \item \textit{Object:} Exoskeleton
            \item \textit{Context:} Reducing muscular load on knee extensor muscles for elderly...
        \end{itemize}
    \end{quote}
\end{itemize}

\textbf{3. Forge (Idea Generator)}
\begin{itemize}
    \item \textbf{Purpose:} To act as the primary AI brainstorming engine, mimicking the cognitive diversity of a human brainstorming group.
    \item \textbf{Input:} The structured AI3C Problem Statement.
    \item \textbf{Process:} Forge is a two-agent sub-system:
    \begin{itemize}
        \item \textbf{Formulator:} A grounded LLM (lower temperature setting) that generates practical, feasible, and incremental ideas.
        \item \textbf{Explorer:} A "wild" LLM (higher temperature setting) that generates out-of-the-box, unconventional, and high-risk/high-reward ideas.
    \end{itemize}
    \item \textbf{Output:} A set of 10 new ideas (5 from each sub-agent) in AOC format, which are appended to the `Idea Vault` alongside the human's ideas from `Muse`.
    \item \textbf{Example:}
    \begin{quote}
        \textbf{Explorer Output:}
        \begin{itemize}
            \item \textit{Idea:} Balloon Cloud Chair
            \item \textit{Action:} Inflates and deflates rhythmically to gently lift or lower...
            \item \textit{Object:} Chair-shaped cluster of responsive balloons
            \item \textit{Context:} Elderly individuals can rise or sit as if buoyed by a cloud...
        \end{itemize}
        \textbf{Formulator Output:}
        \begin{itemize}
            \item \textit{Idea:} Removable Push-Up Cushion
            \item \textit{Action:} Providing extra firmness and spring-back when needed.
            \item \textit{Object:} Seat Cushion
            \item \textit{Context:} Elderly users placing or removing a supportive cushion...
        \end{itemize}
    \end{quote}
\end{itemize}

\subsubsection{Phase 3: Internal Assessment and External Benchmarking}
\label{sec:agent_eval_bench}
\textbf{4. Gatekeeper (Idea Evaluator - Local Novelty Diversity)}
\begin{itemize}
    \item \textbf{Purpose:} To perform the first stage of the 'CA' (Continuous Assessment) pipeline, ensuring "idea abundance" does not become "idea overload." It assesses {local} novelty and diversity.
    \item \textbf{Input:} All unevaluated AOC ideas from `Muse` and `Forge` in the `Idea Vault`.
    \item \textbf{Process:} Gatekeeper uses a semantic characterization framework from an earlier work \cite{Sankar2025b}. It converts all ideas into vector embeddings, computes their cosine similarity, and performs clustering (UMAP + DBSCAN) to identify and shortlist only the most semantically unique and diverse ideas.
    \item \textbf{Output:} A filtered, high-quality list of shortlisted AOC ideas.
\end{itemize}

\textbf{5. Librarian (Research Solutions)}
\begin{itemize}
    \item \textbf{Purpose:} To ground the ideation process in reality by gathering data on existing real-world solutions. This prevents the system from "reinventing the wheel."
    \item \textbf{Input:} The AI3C Problem Statement. It also features a manual mode that allows designers to input specific links or known products.
    \item \textbf{Process:} Uses a web-search-grounded LLM to scan patents, academic papers, and commercial product databases for relevant existing solutions.
    \item \textbf{Output:} A list of existing solutions, structured in AOC format (for direct comparison) and saved to the `Literature Vault`, with source links for credibility.
    \item \textbf{Example:}
    \begin{quote}
        \textbf{Output (AOC Format):}
        \begin{itemize}
            \item \textit{Idea:} SitnStand Portable Smart Rising Seat
            \item \textit{Action:} Inflates and deflates via simple controls to gently lift...
            \item \textit{Object:} Battery-powered inflatable seat cushion
            \item \textit{Context:} Home and outdoor use for elderly...
            \item \textit{Source:} https://www.sitnstand.com
        \end{itemize}
    \end{quote}
\end{itemize}

\textbf{6. Challenger (Idea Evaluator - Global Novelty Diversity)}
\begin{itemize}
    \item \textbf{Purpose:} To perform the most critical evaluation: checking for {global} novelty. An idea is only truly novel if it is different from what already exists.
    \item \textbf{Input:} The shortlisted ideas from `Gatekeeper` (from `Idea Vault`) and the existing ideas from `Librarian` (from `Literature Vault`).
    \item \textbf{Process:} Performs a semantic cross-comparison. It calculates the similarity between each generated idea and the entire corpus of existing solutions. Any idea that is too semantically close to an existing product is filtered out.
    \item \textbf{Output:} A final, globally and locally novel and diverse set of AOC ideas.
\end{itemize}

\subsubsection{Phase 4: Divergence and Feasibility}
\label{sec:agent_div_feas}
\textbf{7. Mint (Action-Object Idea Generator)}
\begin{itemize}
    \item \textbf{Purpose:} To facilitate the 'Diverge' phase and break design fixation. It operationalizes the "ideas from ideas" concept by deconstructing solutions into their core components.
    \item \textbf{Input:} The globally and locally novel AOC ideas from `Challenger`.
    \item \textbf{Process:} Parses the AOC ideas and extracts their fundamental building blocks: a list of `Actions` and a separate list of `Objects`.
    \item \textbf{Output:} Two independent lists (e.g., 20 Actions and 20 Objects), which serve as new creative "seeds."
    \item \textbf{Example:}
    \begin{quote}
        \textbf{Actions:} 1. Elevates and stabilizes... 2. Guides and encourages... 3. Supports and cushions...
        \textbf{Objects:} 1. Adjustable armrests... 2. Ergonomic transition mat... 3. Flexible base platform...
    \end{quote}
\end{itemize}

\textbf{8. Scout (Feasibility Check)}
\begin{itemize}
    \item \textbf{Purpose:} The second 'CA' step, acting as a "combinatorial filter" to manage the creative explosion from `Mint`.
    \item \textbf{Input:} The independent lists of Actions and Objects from `Mint` and the AI3C problem context.
    \item \textbf{Process:} First, it generates all possible `Action-Object` pairs (e.g., 20 x 20 = 400 pairs). Second, it uses an LLM to assign a "feasibility score" (e.g., 1/10) to each pair, evaluating how well that specific combination addresses the problem context.
    \item \textbf{Output:} A sorted list of the highest-scoring, most feasible Action-Object pairs.
    \item \textbf{Example:}
    \begin{quote}
        \textbf{Output:}
        \begin{itemize}
            \item 1. Transforms and adapts to the user’s weight... - Non-slip texture surface — Score: 10/10
            \item 2. Engages and retracts to offer adjustable support... - Reinforced seat back — Score: 10/10
        \end{itemize}
    \end{quote}
\end{itemize}

\subsubsection{Phase 5: Progressive Refinement and Curation}
\label{sec:agent_refine_curate}
\textbf{9. Navigator (Idea Generator)}
\begin{itemize}
    \item \textbf{Purpose:} A second, more refined generation phase that implements "ideas from ideas."
    \item \textbf{Input:} The top-ranked feasible Action-Object pairs from `Scout` and the AI3C problem context.
    \item \textbf{Process:} "Re-hydrates" the abstract Action-Object pairs back into fully-formed, descriptive ideas in the AOC format. This step generates new solutions that are a novel synthesis of the best parts of previous ideas. It then runs its own internal novelty/diversity check.
    \item \textbf{Output:} A new, highly refined set of shortlisted AOC ideas.
\end{itemize}

\textbf{10. Sentinel (Relevance Specialist)}
\begin{itemize}
    \item \textbf{Purpose:} The final 'Curation' phase. It serves as a quality assurance check to ensure that the refined ideas remain perfectly aligned with the original problem statement.
    \item \textbf{Input:} The refined ideas from `Navigator` and `Challenger`, alongside the original problem statement from `Scribe`. Additionally, the designer is free to select and input ideas at this stage.
    \item \textbf{Process:} Performs a rigorous relevance check, comparing each idea's `Context` against the problem's `Criteria` and `Constraints`. It polishes the language and filters any idea that has drifted too far from the original goal.
    \item \textbf{Output:} The final, validated, refined, and relevant list of AOC ideas.
\end{itemize}

\subsubsection{Phase 6: Conceptualization and Visualization}
\label{sec:agent_concept_viz}
\textbf{11. Director (Concept Generator)}
\begin{itemize}
    \item \textbf{Purpose:} To translate the abstract, curated "ideas" into concrete, implementation-ready "concepts."
    \item \textbf{Input:} The final AOC ideas from `Sentinel` and the AI3C problem context.
    \item \textbf{Process:} Expands the simple AOC structure into the rich, detailed \textbf{PFIC (Principle, Features, Implementation, Characteristics)} format \cite{Sankar2025a}.
    \item \textbf{Output:} A set of structured PFIC concepts saved to the `Concept Vault`.
    \item \textbf{Example:}
    \begin{quote}
        \textbf{Input Idea:} Integrated Safety Belt with Retractable Assistance
         
        \textbf{Output (PFIC Format):}
        \begin{itemize}
            \item \textit{Principle:} Providing adjustable mechanical support through a retractable safety belt...
            \item \textit{Features:} Retractable safety belt, Adjustable tension control, Automatic engagement...
            \item \textit{Implementation:} Incorporate a motorized retracting mechanism... Use sensors to detect user movement...
            \item \textit{Characteristics:} Safe and reliable support, User-friendly automatic operation...
        \end{itemize}
    \end{quote}
\end{itemize}

\textbf{12. Leo (Image Generator)}
\begin{itemize}
    \item \textbf{Purpose:} The final 'Conceptualize' step, providing photorealistic visualizations of the concepts.
    \item \textbf{Input:} The PFIC structured concepts from `Director`.
    \item \textbf{Process:} Uses the `Features` and `Characteristics` from the PFIC format as a detailed prompt for a text-to-image generation model.
    \item \textbf{Output:} A set of photorealistic product renderings that visually represent the final concepts.
\end{itemize}

\subsection{Technical Implementation}
\label{sec:midas_technical}

The MIDAS framework is not just a conceptual model; it is a functioning technical system built on a "compound LLM" approach, as detailed in Figure \ref{fig:midas_tech_impl}.

\subsubsection{A Compound LLM Approach}
\label{sec:midas_compound_llm}
\begin{figure*}[t!]
    \centering
    \includegraphics[width=0.85\linewidth]{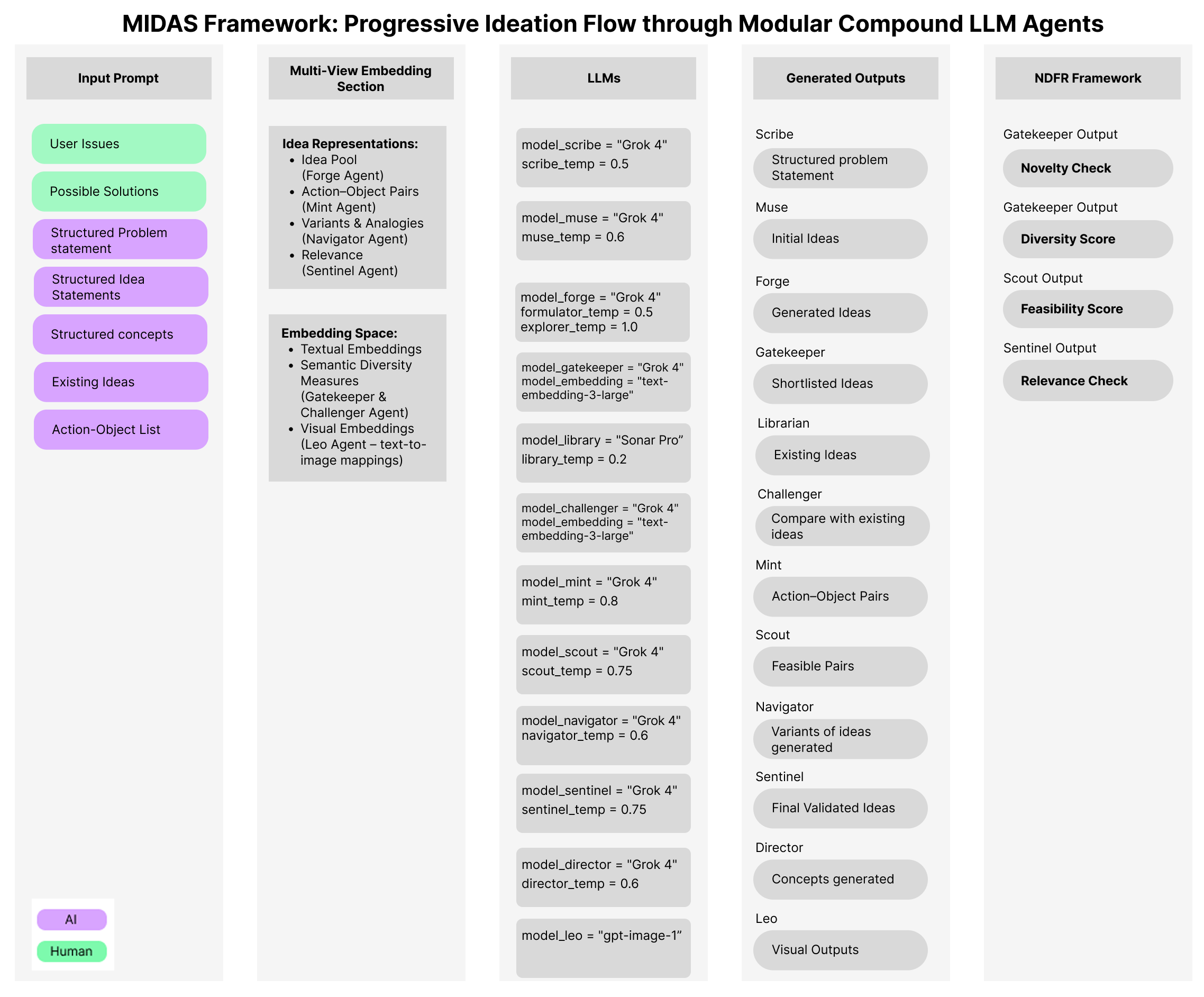} 
    \caption{Technical Implementation Flowchart of the MIDAS Framework, showing Input Prompts, Multi-View Embedding, specific LLM assignments for each agent, Generated Outputs, and the NDFR Evaluation Framework.}
    \label{fig:midas_tech_impl}
\end{figure*}

Instead of relying on a single, general-purpose LLM, MIDAS employs an ensemble of models, selecting the "right tool for the right job." This agentic approach allows for fine-grained control over the creative process. As shown in Figure \ref{fig:midas_tech_impl}, different agents utilize different models and parameter settings:
\begin{itemize}
    \item \textbf{Generative \& Reasoning Tasks} (e.g., `Scribe`, `Forge Formulator`, `Director`, `Navigator`) may use a powerful, coherent model like `Grok 4 Reasoning` with a lower temperature (e.g., `temp=0.5` or `0.6`) for precision.
    \item \textbf{Creative Exploration} (e.g., `Forge Explorer`) uses the same model but with a high temperature (e.g., `temp=1.0`) to encourage "wild" and diverse outputs.
    \item \textbf{Search \& Grounding} (e.g., `Librarian`) uses a model with integrated web-search capabilities (e.g., `Sonar Pro` with `temp=0.2`) to fetch real-world data.
    \item \textbf{Image Generation} (e.g., `Leo`) uses a dedicated text-to-image model (e.g., `gpt-image-1`).
\end{itemize}
This compound approach provides a level of control and specialization that is impossible to achieve with a single LLM.

\subsubsection{Embedding for Semantic Evaluation}
\label{sec:midas_embedding}
The 'CA' (Continuous Assessment) pipeline, integral to the `Gatekeeper`, `Challenger`, and `Scout` agents, is powered by a multi-view embedding system. This system translates textual and even visual concepts into a high-dimensional vector space, allowing for mathematical computation of their creative qualities.
\begin{itemize}
    \item \textbf{Idea Representations:} The textual AOC outputs from `Forge` and `Muse` are converted into text embeddings.
    \item \textbf{Semantic Similarity:} We primarily use Cosine Similarity to measure the "distance" between two idea vectors. A large distance implies high \textit{novelty}.
    \item \textbf{Clustering \& Diversity:} To measure \textit{diversity}, we run clustering algorithms (UMAP for dimensionality reduction, DBSCAN for cluster identification) on the entire idea pool. A pool with many small, distinct clusters is considered more diverse. This is the quantitative backbone of our 'NDFR' (Novelty, Diversity, Feasibility, Relevance) framework, based on the semantic characterization methods established in \cite{Sankar2025b}.
\end{itemize}
This computational approach to evaluation is what enables the 'CG/CA' loop, allowing the system to progressively refine its own outputs towards a truly novel and diverse set of final concepts.

\subsection{Evaluation Framework: The NDFR Criteria}
\label{sec:methodology_ndfr}

The claim of "better ideas" is subjective and ambiguous. To introduce analytical rigour, we posit that the success of an ideation system cannot be measured by a single metric (e.g., "creativity" or "quantity"). We therefore integrate a multi-dimensional evaluation heuristic in MIDAS: the \textbf{NDFR Framework}. This framework assesses the idea across four criteria: Novelty, Diversity, Feasibility, and Relevance.

\subsubsection{Novelty}
We argue that "novelty" in design ideation must be bifurcated into two distinct types:
\begin{itemize}
    \item \textbf{Local Novelty:} This measures the originality of an idea relative to the other ideas generated within the same session. A system that produces 1000 ideas, all of which are minor variations of each other, has very low local novelty. This metric, computationally assessed by the `Gatekeeper` agent, quantifies the system's ability to avoid self-referential fixation and semantic variants.
    \item \textbf{Global Novelty:} This is the true measure of innovation. It assesses the originality of an idea relative to the entire corpus of existing, real-world solutions (e.g., patents, academic papers, and commercial products). An idea can be locally novel but globally trivial (i.e., "reinventing the wheel"). This criterion is computationally managed by the `Challenger` agent and provides a grounding in reality, filtering for genuine, state-of-the-art innovation.
\end{itemize}

\subsubsection{Diversity}
Distinct from novelty (which can apply to a single idea), diversity is a property of the entire idea space. It measures the semantic "spread" or breadth of the generated solutions. A high-diversity output will explore multiple, disparate conceptual families. For instance, in solving the problem of a chair for the elderly, a high-diversity output might include mechanical solutions, biometric solutions, inflatable solutions, and material-science solutions. A low-diversity output might only provide 50 variations of a mechanical spring. We quantify this using the vector-based clustering methods established in a prior work \cite{Sankar2025b}, measuring metrics such as "idea sparsity" and "cluster sparsity."

\subsubsection{Feasibility}
An idea that is novel and diverse becomes a useless concept if it is pure fantasy. Feasibility assesses the practical and technical viability of an idea when it turns into a concept. It asks: Can this be built with known technology and resources? Does it adhere to the fundamental constraints (e.g., physics, cost, ergonomics) of the problem? This criterion is explicitly handled by the `Scout` agent (assessing action-object pairings) and is a core component of the `PFIC` structure generated by the `Director` agent, specifically in the `Implementation` and `Characteristics` fields.

\subsubsection{Relevance}
Finally, an idea can be novel, diverse, and feasible, yet fail to solve the actual problem. Relevance measures the alignment between a proposed concept and the original problem statement. It assesses how well the solution directly addresses the core `Contradiction`, `Criteria`, and `Constraints` defined in the structured `AI3C` problem format. This goal-alignment check is the primary function of the `Sentinel` agent, ensuring that the progressive pipeline does not drift from the user's core need.

Together, these four NDFR criteria embedded in the MIDAS system provide a comprehensive, rigorous, and computationally assessable framework to move beyond subjective claims and empirically evaluate the output of the idea generated by the generator agents.

\section{Research Questions and Methodology}
\label{sec:methodology}

The development of the MIDAS, as detailed in Section \ref{sec:midas_framework}, is predicated on a central hypothesis: that a distributed, agentic, and progressive approach to ideation is superior to the prevailing single-step, monolithic LLM paradigm.

\subsection{Research Question (R.Q.)}
\label{sec:methodology_rq}

Our primary investigation is encapsulated in one overarching research question (R.Q.).

\textbf{R.Q.: Can an Agentic AI-based Progressive Ideation System (MIDAS) produce a set of ideas that are demonstrably more novel, diverse, and feasible compared to single-stage, non-agentic AI ideation methods?}

To comprehensively answer this, we decompose this primary question into several specific lines of inquiry that our methodology must address:
\begin{enumerate}
    \item \textbf{Efficacy of the Pipeline:} Does the progressive, multi-stage 'CG/CA' (Continuous Generation / Continuous Assessment) pipeline within MIDAS result in a final set of concepts with quantifiably higher quality (as measured by our NDFR criteria) than the "idea abundance" generated by a single-stage, generalist LLM interaction?
    \item \textbf{Quality of Novelty:} Does the integrated global novelty check (via the `Librarian` and `Challenger` agents) successfully filter for "true" innovation, producing ideas that are novel against real-world solutions, rather than just semantic variants?
    \item \textbf{Human-Centric Collaboration:} How does the 'PAC' (Participatory, Active, Collaborative) model, which positions the human as an active idea contributor via the `Muse` agent, impact the designer's perceived engagement, sense of ownership, and the overall fluency of the co-creative experience?
\end{enumerate}
Answering these questions requires a robust demonstration as detailed below.

\section{Demonstrating Progressive Ideation through Multi-Domain Validation}
\label{sec:midas_demonstration}

To validate the scalability and consistency of the MIDAS, we conducted a comprehensive series of demonstrations across six distinct design problems (PS1 to PS6). The core objective was to ascertain whether the agentic, progressive pipeline could consistently yield a curated set of truly novel and diverse ideas that satisfy the stringent \textit{NDFR (Novelty, Diversity, Feasibility, Relevance)} criteria while maintaining a high level of human-AI collaborative synergy.

In this demonstration, a group of six novice designers (M.Des students) utilized the MIDAS platform to tackle the six problems (PS1 to PS6). Unlike the earlier single-spurt abundance experiments where designers spent hours sifting through clusters of similar ideas, each designer in this study engaged with MIDAS for a maximum of \textit{20 minutes per session}. Adhering to the PAC model, the process began with the designer defining the problem context through the \textit{Scribe} agent. Crucially, the designers did not remain passive spectators; instead, they actively seeded the process by contributing a handful of their own initial thoughts into the \textit{Muse} agent, which were then seamlessly integrated into the computational pipeline.

As detailed in Table \ref{tab:agent_filtering_metrics}, the MIDAS employs a rigorous "filtering and refinement" architecture, and the table shows the number of ideas that passed through each agent for each of the six problem statements. While the \textit{Forge} agent initially generated a significant volume of raw ideas (ranging from 50 to 80 per session), which was subsequently reduced by \textit{Gatekeeper} based on \textit{local novelty}. The subsequent intervention of the \textit{Librarian} and \textit{Challenger} agents ensured that only those concepts demonstrating genuine \textit{global novelty} relative to existing literature and patents were permitted to proceed. The final curated idea sets (Sentinel output) were notably focused, with PS1 (24), PS2 (19), PS3 (19), PS4 (15), PS5 (11), and PS6 (17) being the most prominent. In each case, the AI-generated contributions remained numerically higher than the human-seeded ideas, and the final concepts were often perceived as organic evolutions of the collaborative human-AI dialogue. Further more, based on the ideas generated by forge, the subsequent agents in the pipeline namely \textit{mint}, \textit{scout} and \textit{navigator} were able to source ideas from ideas as proposed earlier. From Table \ref{tab:agent_filtering_metrics}, it is clear that out of the total number of ideas that were finally curated, there was an almost equal distribution of ideas that originated from generating ideas while solving the problem and from generating ideas themselves. Our proposition of generating ideas from ideas worked exceptionally well in the MIDAS framework, proving that there are opportunities for cross-pollination of ideas between different problem statements.

\begin{table}[h!]
\centering
\caption{Quantitative Idea Filtering and Generation Count Metrics across the MIDAS Agent Ensemble.}
\label{tab:agent_filtering_metrics}
\small
\begin{tabular}{lcccccc}
\toprule
\textbf{Agent / Phase} & \textbf{PS1} & \textbf{PS2} & \textbf{PS3} & \textbf{PS4} & \textbf{PS5} & \textbf{PS6} \\
\midrule
Scribe (Problem Def.)          & 1   & 1   & 1   & 1   & 1   & 1   \\
Muse (Human Ideas)             & 8   & 6   & 6   & 5   & 4   & 6   \\
Forge (Raw AI Ideas)           & 75  & 70  & 65  & 60  & 55  & 78  \\
Gatekeeper (Shortlist)         & 32  & 28  & 26  & 22  & 18  & 31  \\
Librarian (Prior Art)          & 10  & 10  & 10  & 10  & 10  & 10  \\
Challenger (Globally Novel)    & 11  & 9   & 8   & 7   & 5   & 8   \\
Mint (A-O Pairs)               & 20  & 20  & 20  & 20  & 20  & 20  \\
Scout (Feasibility Eval)       & 400 & 400 & 400 & 400 & 400 & 400 \\
Navigator (Novel/Feasible)     & 13  & 10  & 11  & 8   & 6   & 9   \\
\textbf{Sentinel (Final Curated)} & \textbf{24} & \textbf{19} & \textbf{19} & \textbf{15} & \textbf{11} & \textbf{17} \\
Director (Concepts)            & 22  & 17  & 18  & 14  & 10  & 15  \\
Leo (Renderings)               & 20  & 15  & 18  & 12  & 9   & 15  \\
\bottomrule
\end{tabular}
\end{table}

\begin{figure*}[ht!]
\centering
\begin{subfigure}[b]{0.4\textwidth}
\centering
\includegraphics[width=\linewidth]{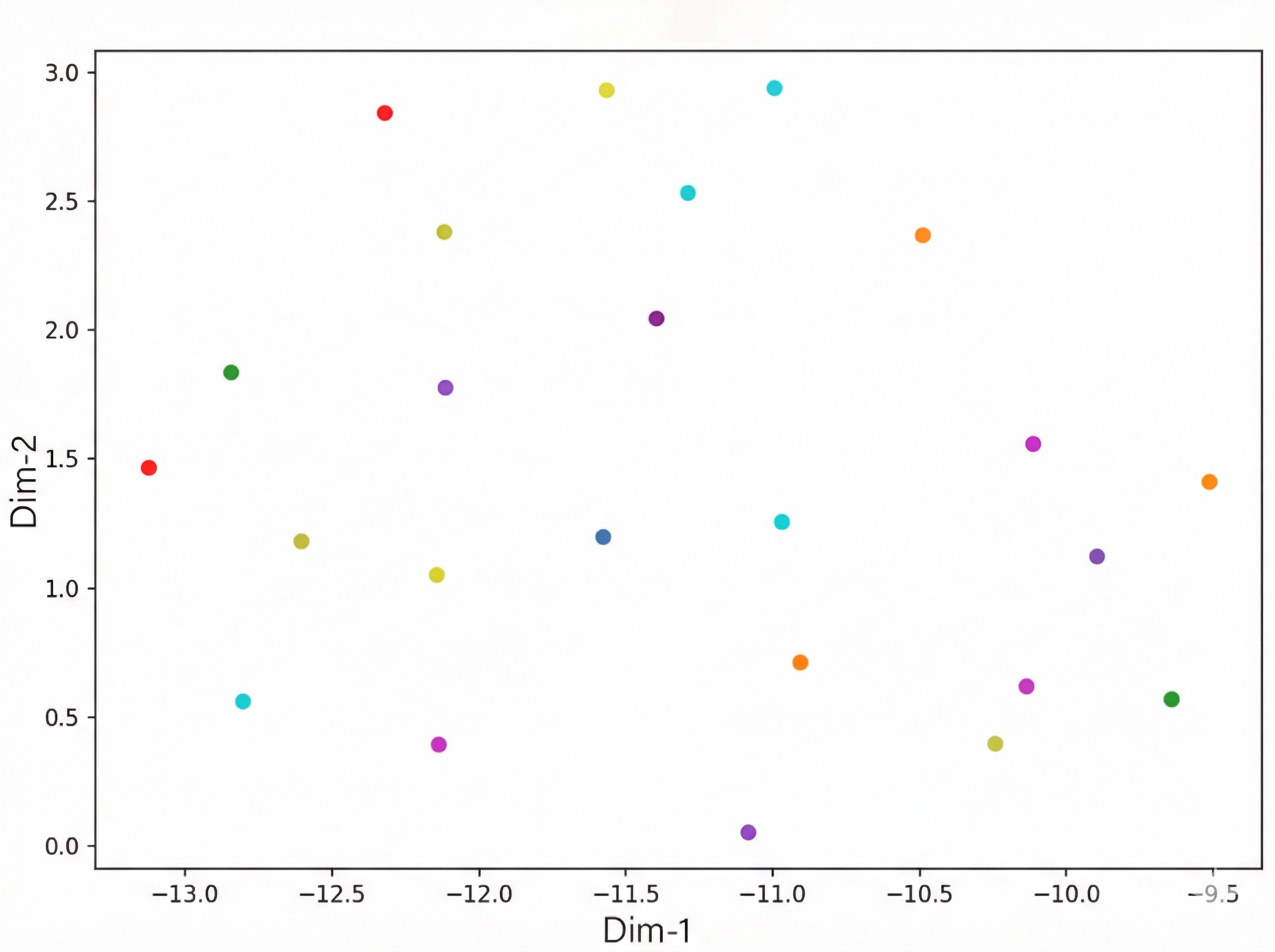}
\caption{PS1: Waste Management (24 Ideas)}
\label{fig:midas_ps1}
\end{subfigure}
\hfill
\begin{subfigure}[b]{0.4\textwidth}
\centering
\includegraphics[width=\linewidth]{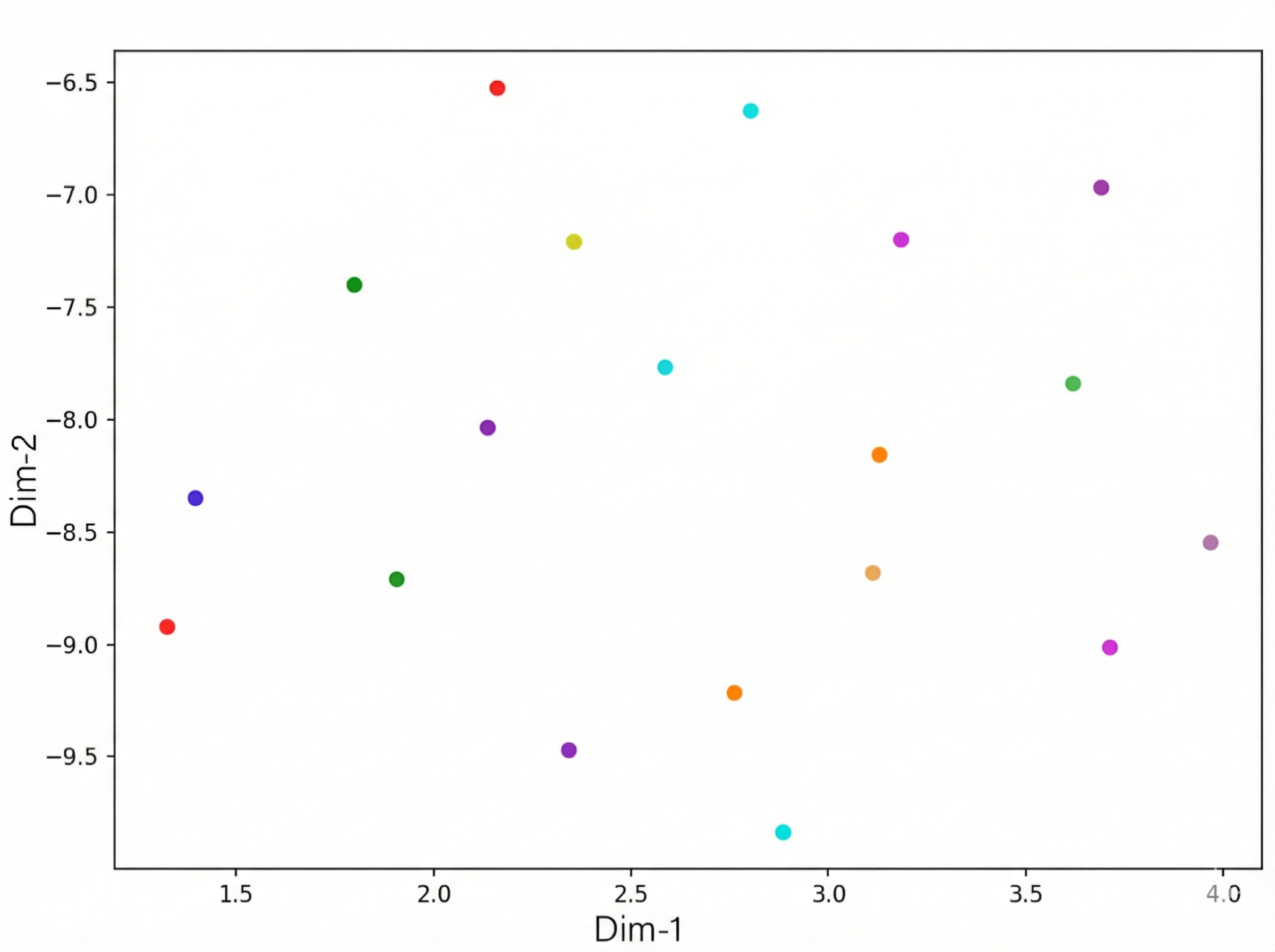}
\caption{PS2: Footwear Disinfection (19 Ideas)}
\label{fig:midas_ps2}
\end{subfigure}

\vspace{0.5cm}

\begin{subfigure}[b]{0.4\textwidth}
\centering
\includegraphics[width=\linewidth]{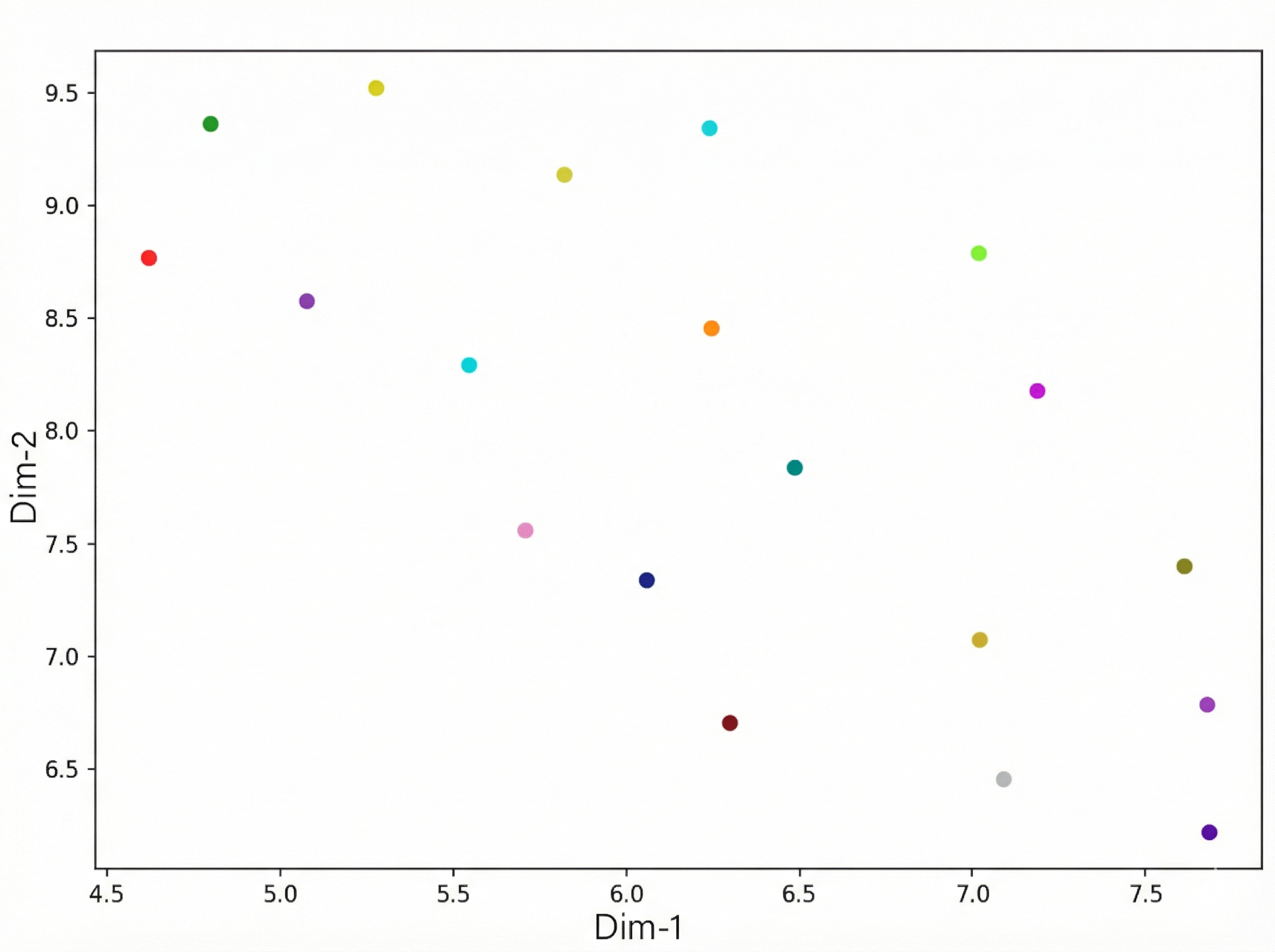}
\caption{PS3: Dish Cleaning (19 Ideas)}
\label{fig:midas_ps3}
\end{subfigure}
\hfill
\begin{subfigure}[b]{0.4\textwidth}
\centering
\includegraphics[width=\linewidth]{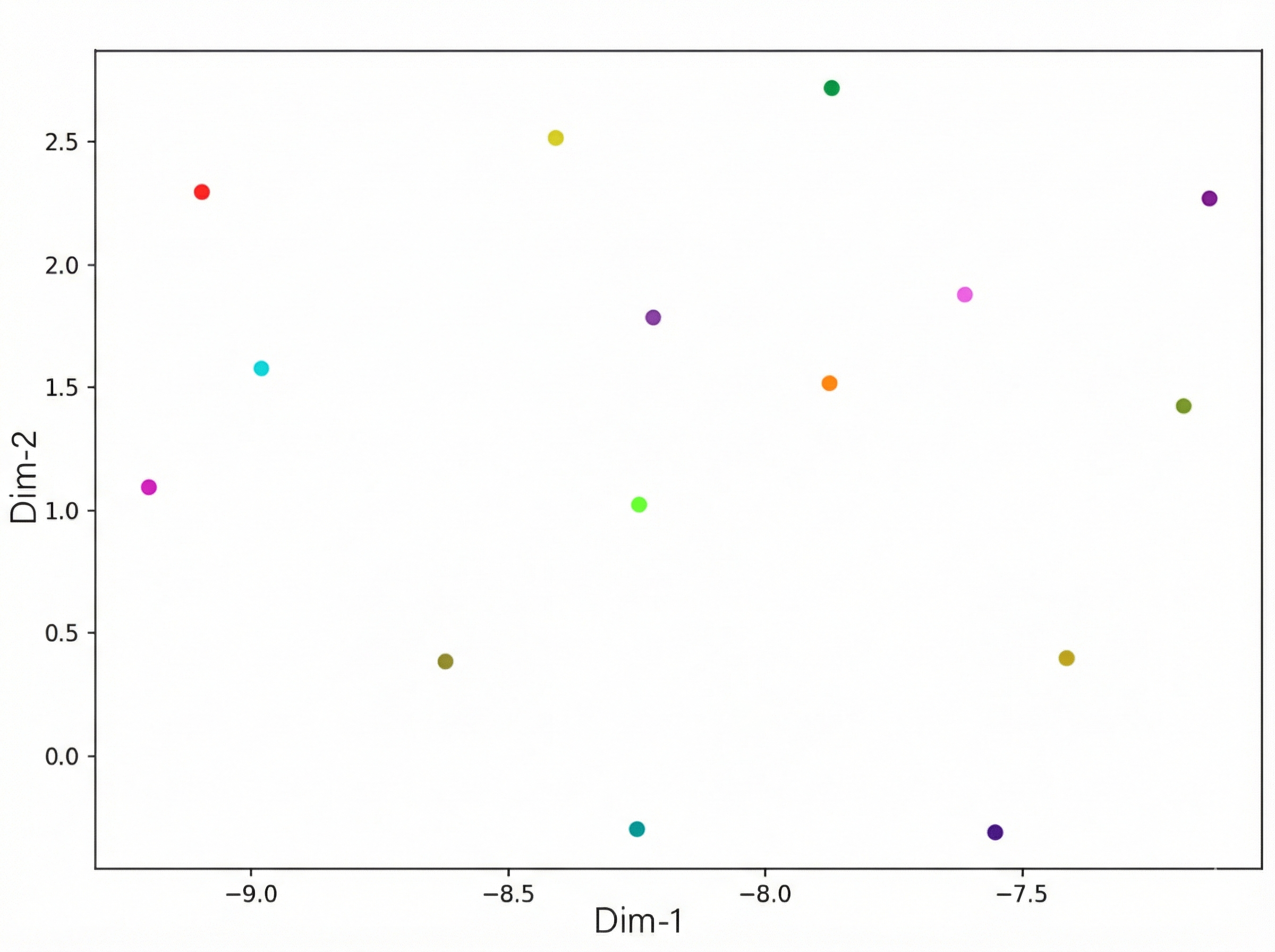}
\caption{PS4: Standing Comfort (15 Ideas)}
\label{fig:midas_ps4}
\end{subfigure}

\vspace{0.5cm}

\begin{subfigure}[b]{0.4\textwidth}
\centering
\includegraphics[width=\linewidth]{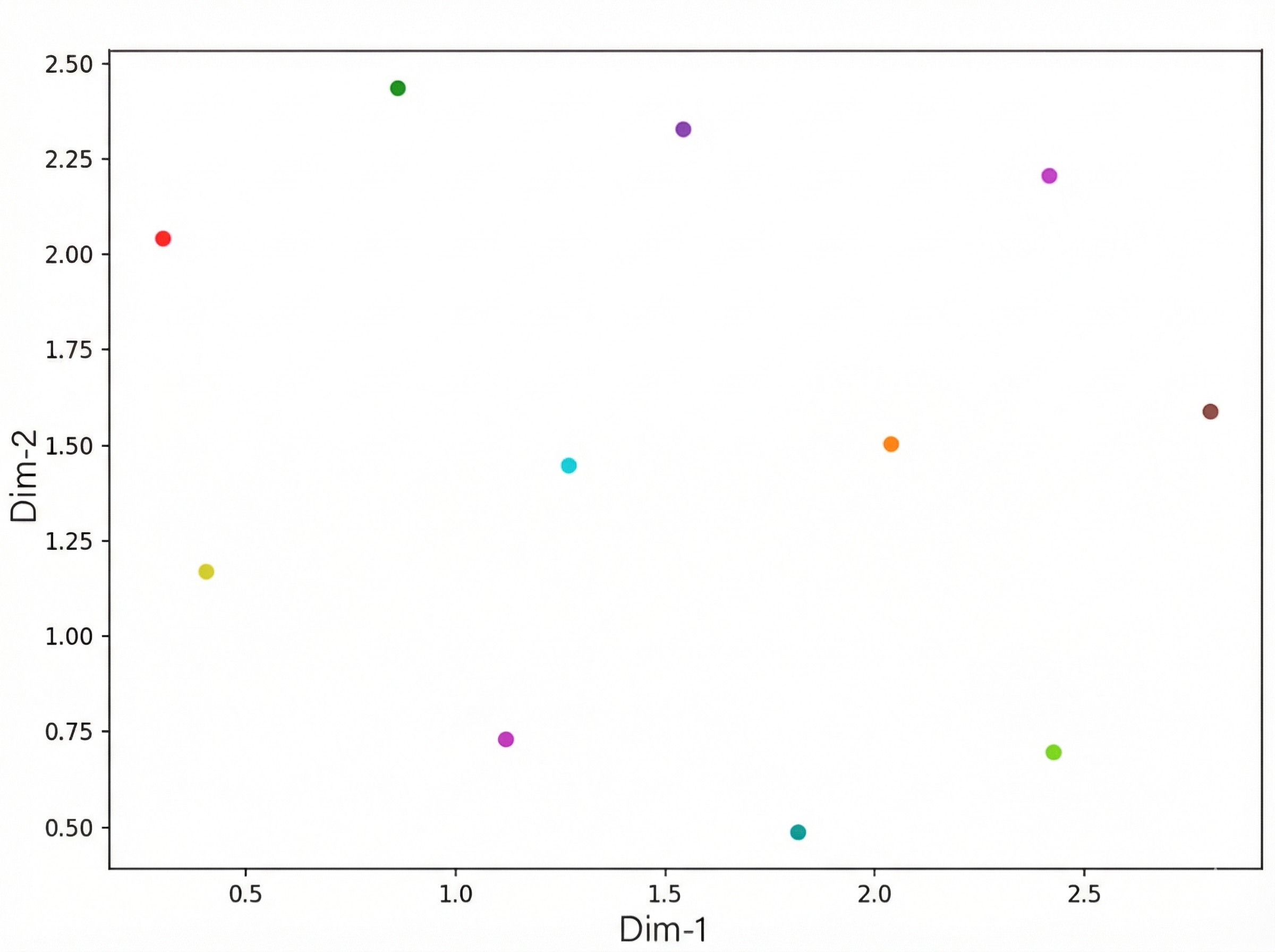}
\caption{PS5: Bird-Feeding (11 Ideas)}
\label{fig:midas_ps5}
\end{subfigure}
\hfill
\begin{subfigure}[b]{0.4\textwidth}
\centering
\includegraphics[width=\linewidth]{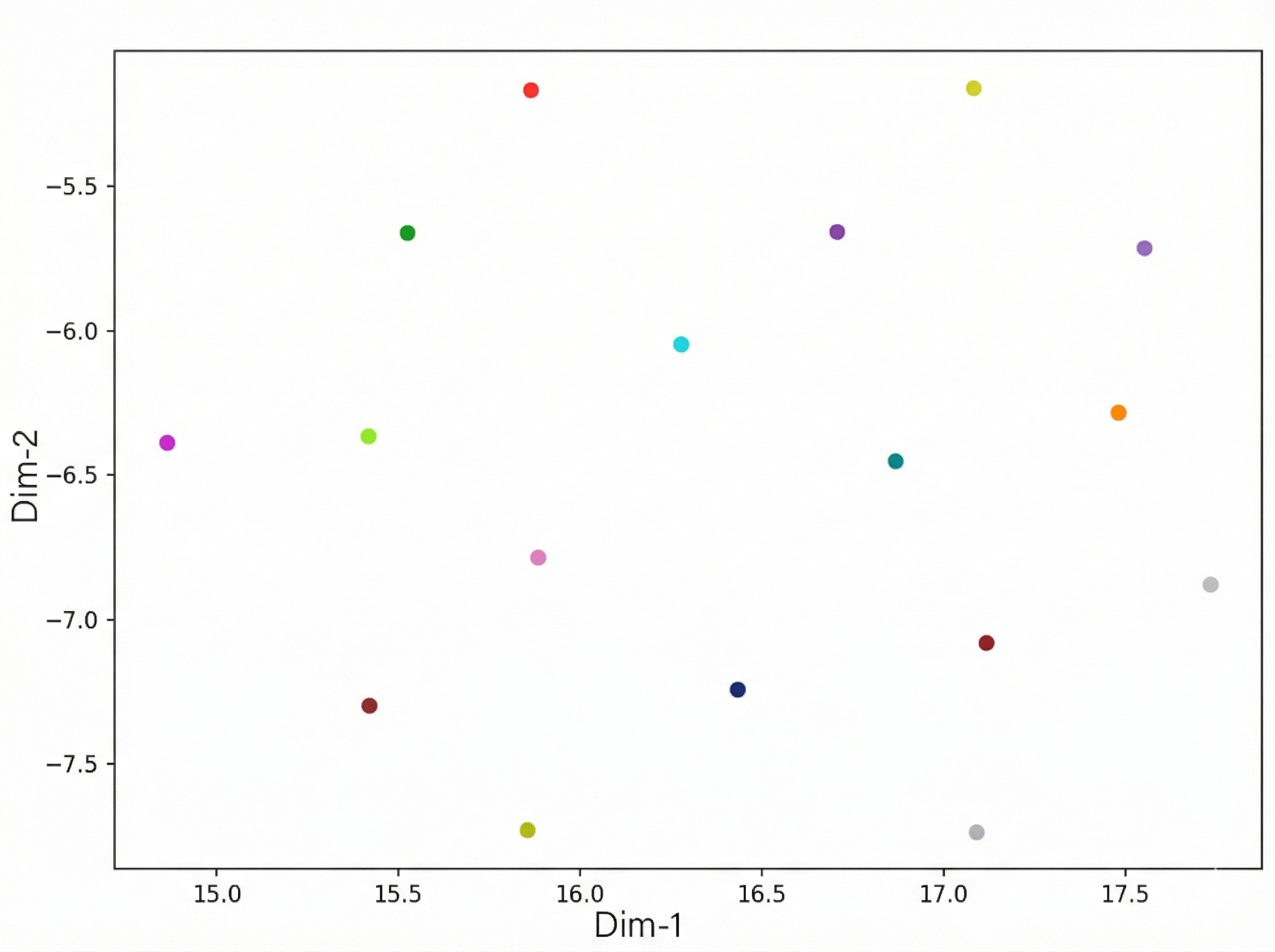}
\caption{PS6: Umbrella Storage (17 Ideas)}
\label{fig:midas_ps6}
\end{subfigure}

\caption{Cluster Plot for ideas across six design problems using the MIDAS framework. Across all problems, the DBSCAN algorithm identifies the entire idea pool as "noise," indicating that the system consistently produces highly diverse, non-overlapping conceptual sets that break the semantic fixation common in monolithic LLM outputs.}
\label{fig:midas_multi_cluster_plots}
\end{figure*}

The semantic quality of these curated outputs was objectively verified using the vector-based characterization framework \cite{Sankar2025b}. As visualized in Figure \ref{fig:midas_multi_cluster_plots}, the ideas generated for each of the six problem statements yielded a cluster scatter plot where the DBSCAN algorithm classified the entire output as a single "noise" cluster. This phenomenon is highly significant in design research, as it implies that every single idea in the final set is so semantically distinct that no two ideas can be grouped together. This provides empirical proof of extreme local novelty and diversity, addressing the primary research inquiry regarding the efficacy of the 'CG/CA' pipeline. While the previous 100-idea single-spurt studies resulted in dense clusters of variants (Figures \ref{fig:results_ps1_ps2}, \ref{fig:results_ps3_ps4}, \ref{fig:results_ps5_ps6}), MIDAS delivered a handful of ideas that are not only distinct from each other but also novel relative to the existing solutions identified by the \textit{Librarian} agent.

Furthermore, the human-centric nature of the collaboration was highlighted by the flexibility of the \textit{Sentinel} and \textit{Director} phases. Designers frequently exercised their agency to further refine, add to, or prune ideas before final conceptualization, resulting in slight variations in the final counts for \textit{Director} and \textit{Leo} (Table \ref{tab:agent_filtering_metrics}). This active intervention underscores the 'PAC' model's success; the AI does not simply 'dump' a result, but serves as a true \textit{Thought Provocation System} that the human designer can steer towards the most promising and relevant directions within a quick 20-minute window.

\begin{figure*}[ht!]
    \centering
    \footnotesize
    \begin{subfigure}[b]{0.31\textwidth}
        \includegraphics[width=\linewidth]{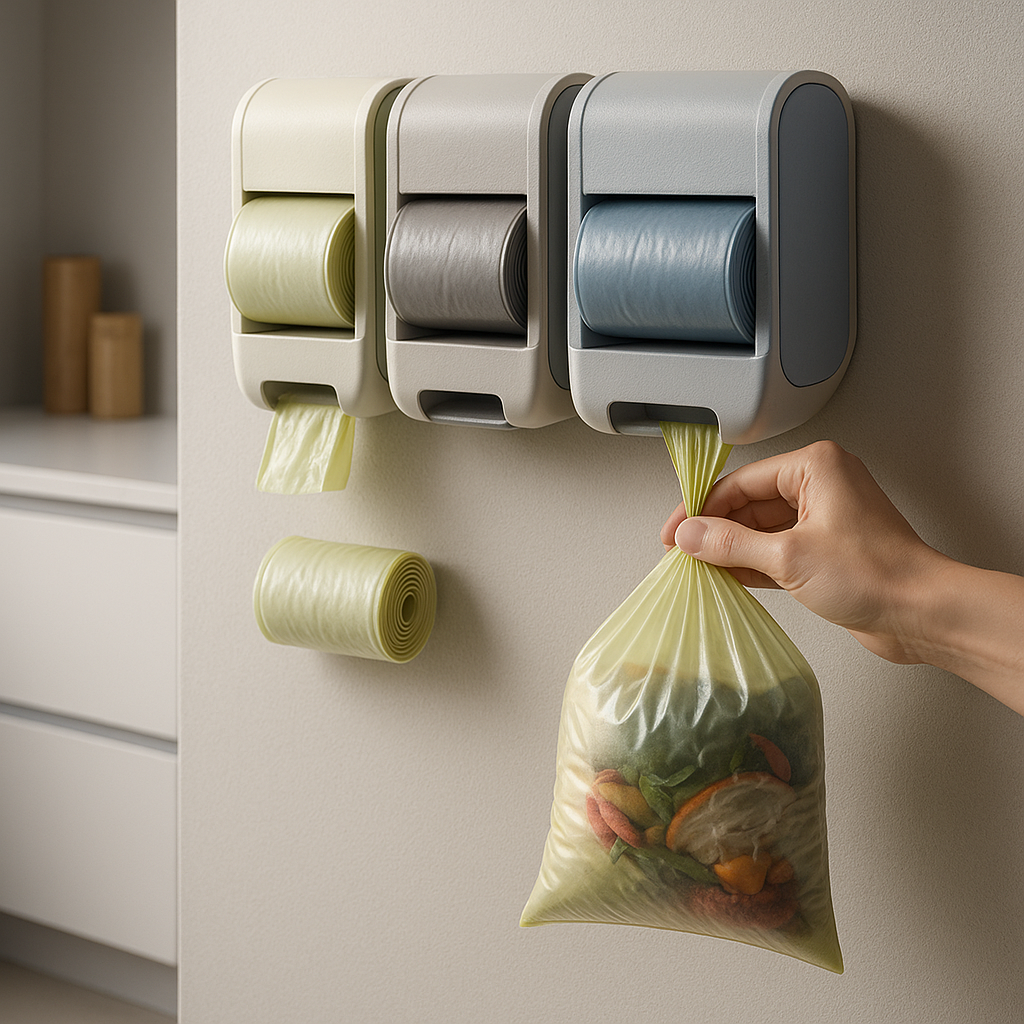}
        \caption{PS1: Collapsible Category Sleeves}
    \end{subfigure}
    \hfill
    \begin{subfigure}[b]{0.31\textwidth}
        \includegraphics[width=\linewidth]{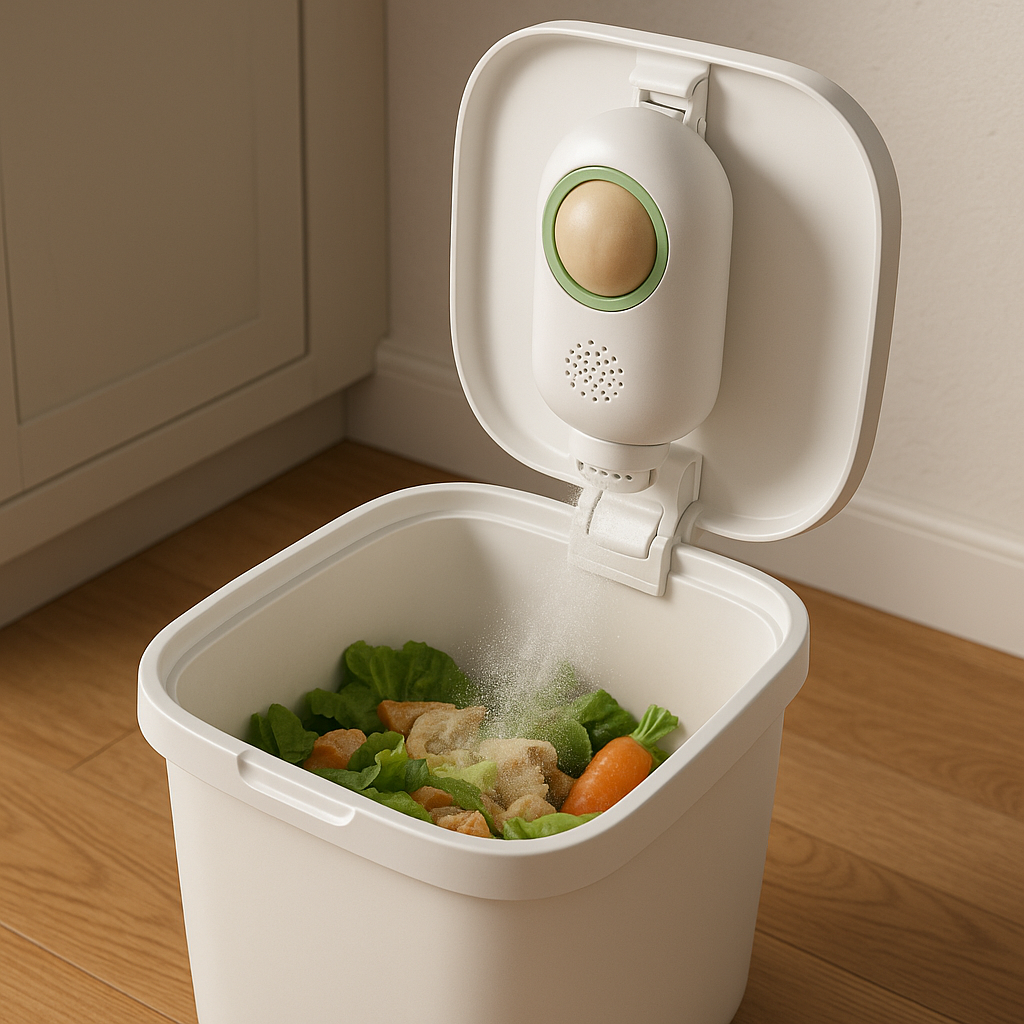}
        \caption{PS1: Natural Enzyme Sprinklers}
    \end{subfigure}
    \hfill
    \begin{subfigure}[b]{0.31\textwidth}
        \includegraphics[width=\linewidth]{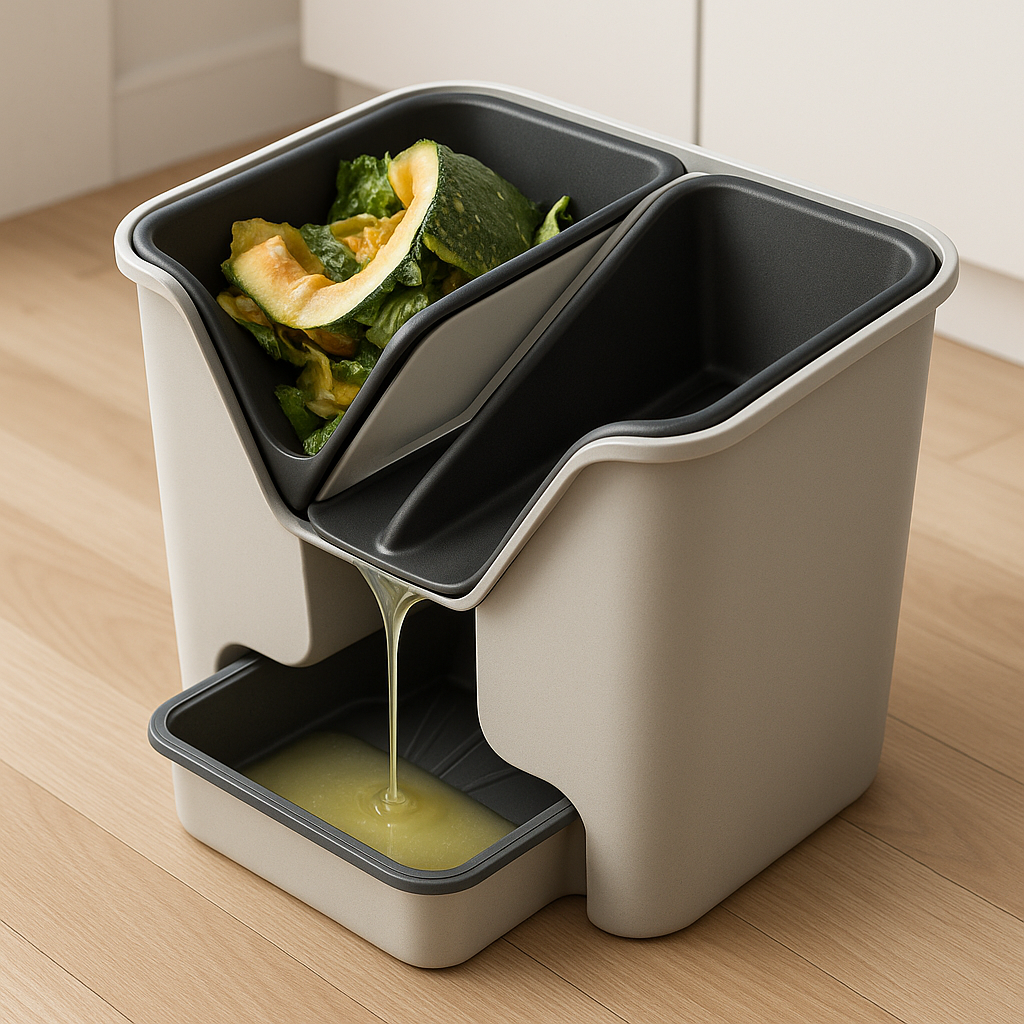}
        \caption{PS1: Leak-Proof Funnel Barriers}
    \end{subfigure}
    \vspace{0.2cm}

    \begin{subfigure}[b]{0.31\textwidth}
        \includegraphics[width=\linewidth]{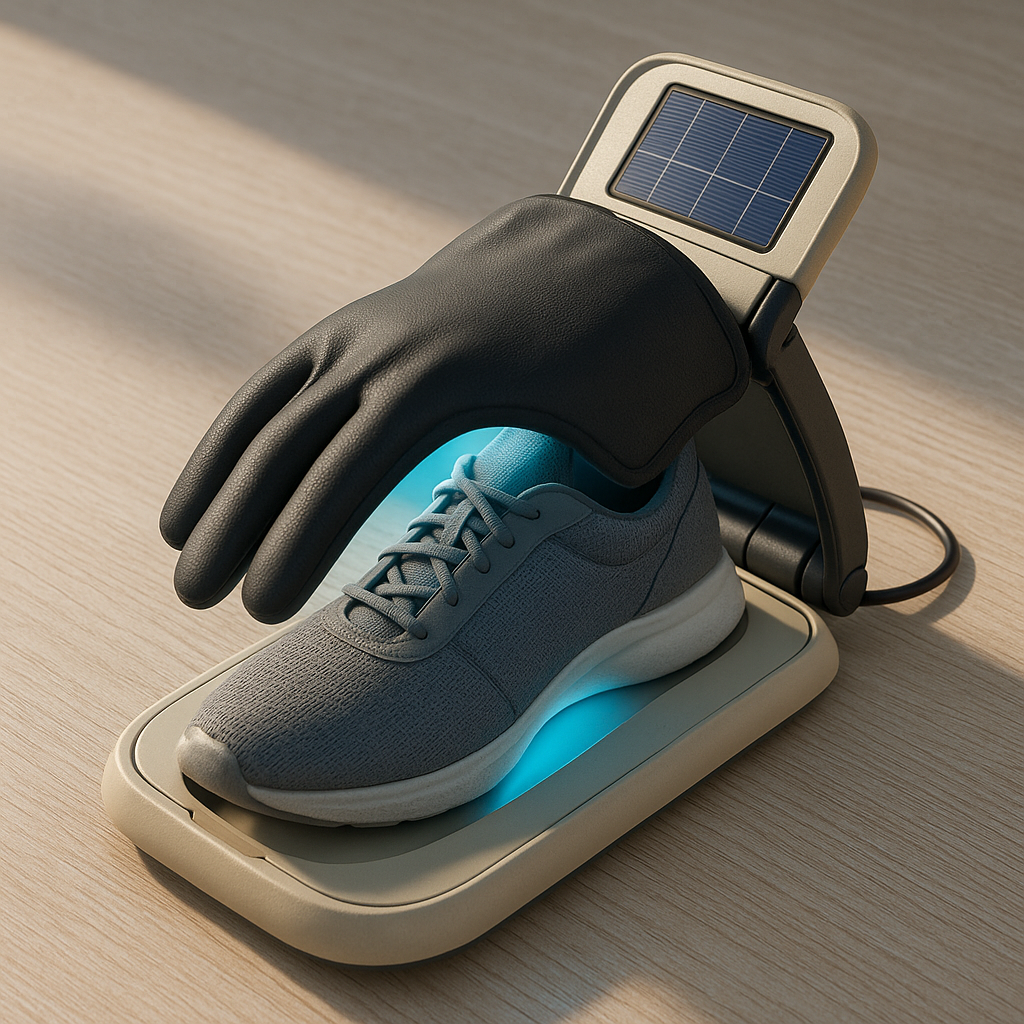}
        \caption{PS2: Solar-Powered UV Glove}
    \end{subfigure}
    \hfill
    \begin{subfigure}[b]{0.31\textwidth}
        \includegraphics[width=\linewidth]{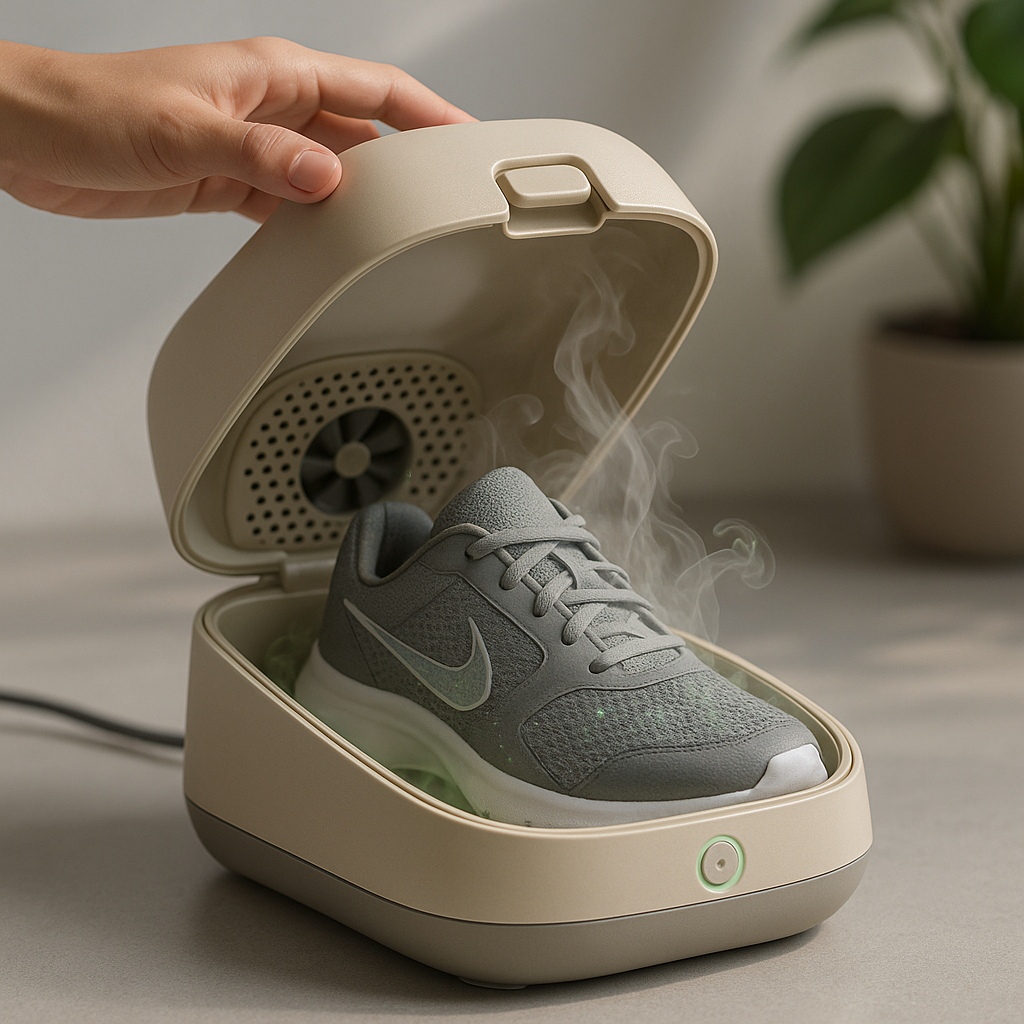}
        \caption{PS2: Purifying Vapor Chamber}
    \end{subfigure}
    \hfill
    \begin{subfigure}[b]{0.31\textwidth}
        \includegraphics[width=\linewidth]{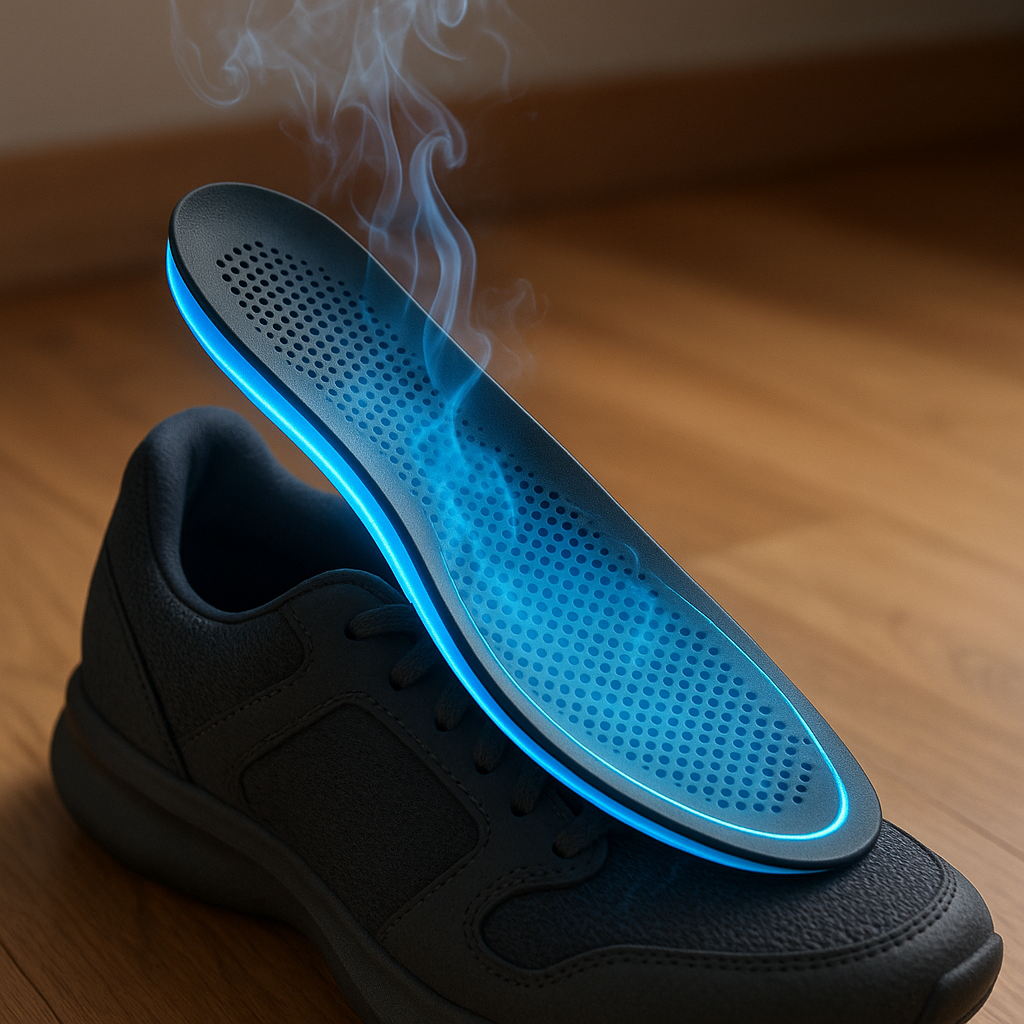}
        \caption{PS2: Photocatalytic Odor Neutralizer}
    \end{subfigure}
    \vspace{0.2cm}

    \begin{subfigure}[b]{0.31\textwidth}
        \includegraphics[width=\linewidth]{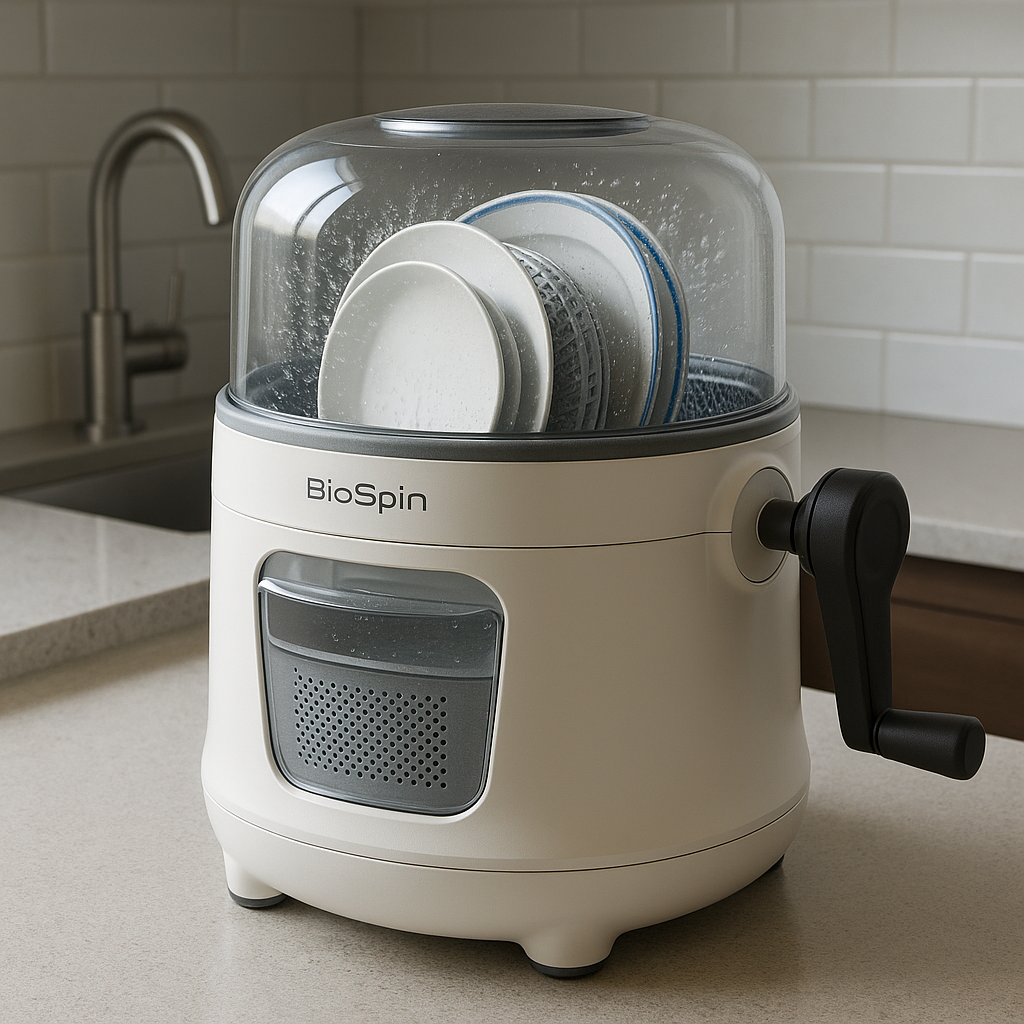}
        \caption{PS3: Bio-Spin Water Recycler}
    \end{subfigure}
    \hfill
    \begin{subfigure}[b]{0.31\textwidth}
        \includegraphics[width=\linewidth]{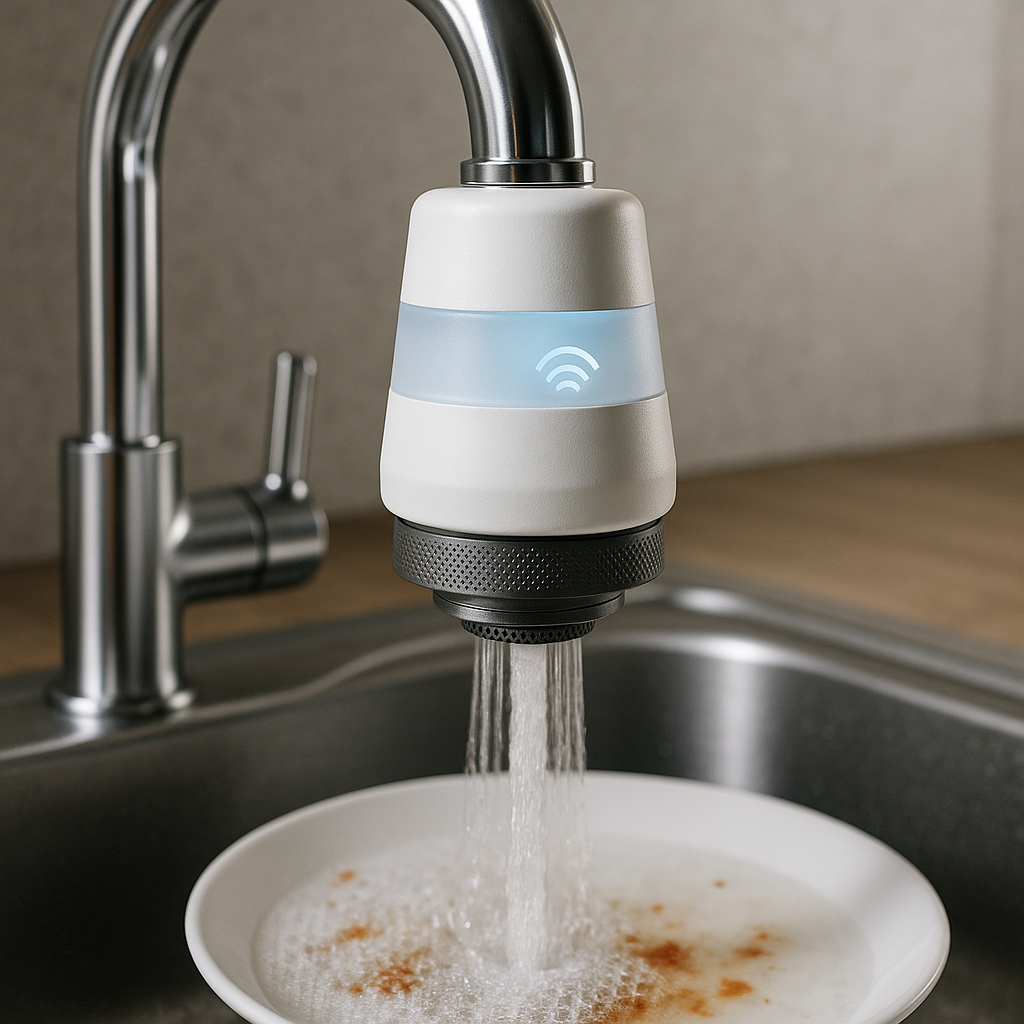}
        \caption{PS3: Smart Faucet Adapter}
    \end{subfigure}
    \hfill
    \begin{subfigure}[b]{0.31\textwidth}
        \includegraphics[width=\linewidth]{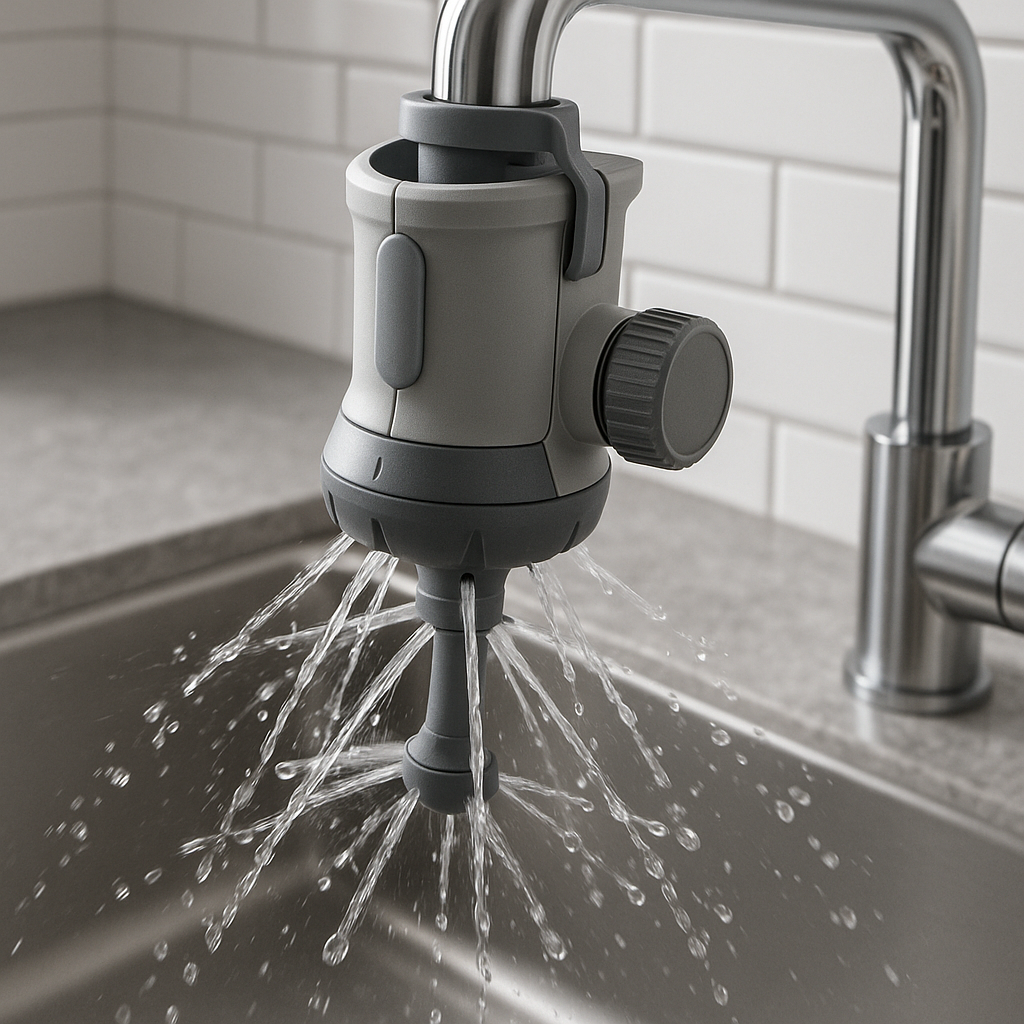}
        \caption{PS3: Countercurrent Flow Nozzle}
    \end{subfigure}

    \caption{Concept renderings generated by the Leo agent for PS1–PS3. These visualizations represent the culmination of the progressive ideation pipeline, where structured PFIC concepts are translated into photorealistic forms to provoke final designer reflections.}
    \label{fig:leo_concept_renders_ps1_ps3}
\end{figure*}

\begin{figure*}[ht!]
    \centering
    \footnotesize
    \begin{subfigure}[b]{0.31\textwidth}
        \includegraphics[width=\linewidth]{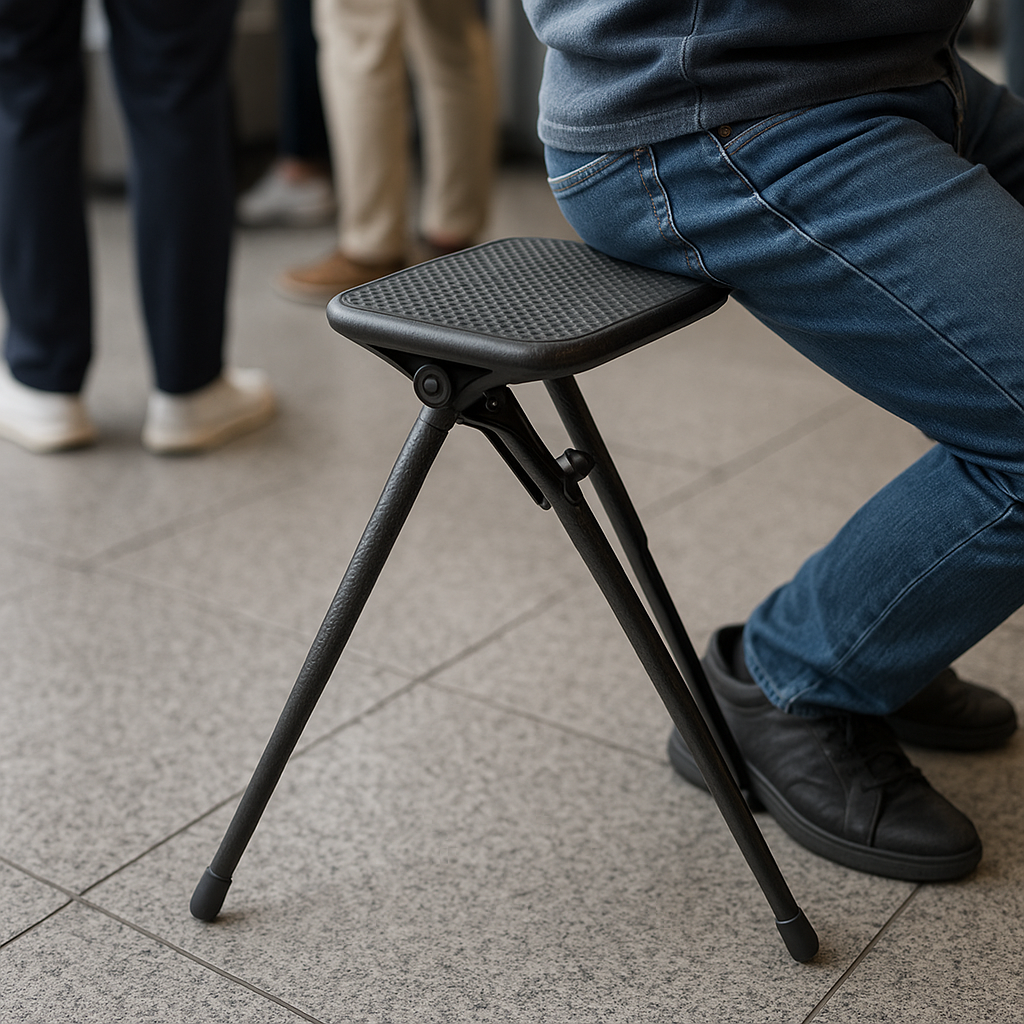}
        \caption{PS4: Foldable Mini Stool}
    \end{subfigure}
    \hfill
    \begin{subfigure}[b]{0.31\textwidth}
        \includegraphics[width=\linewidth]{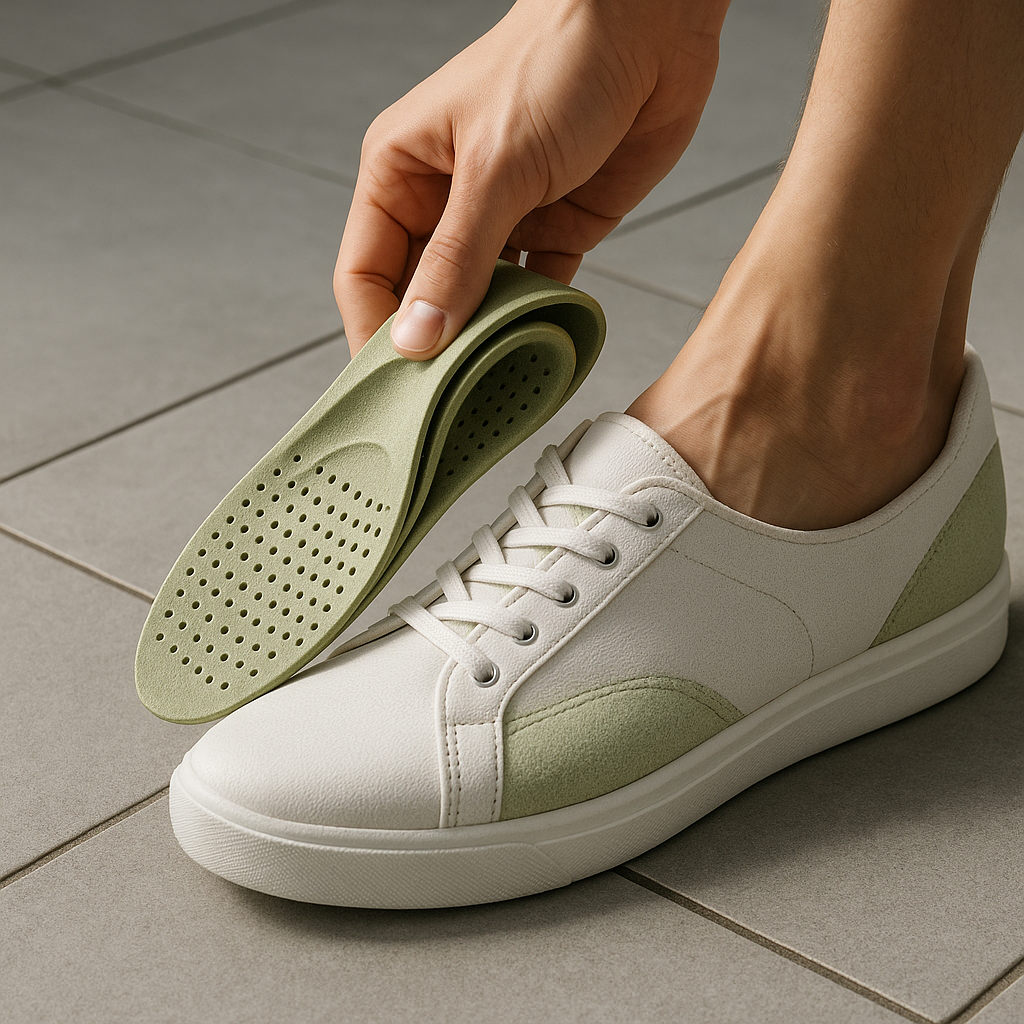}
        \caption{PS4: Eco-Foam Foot Insert}
    \end{subfigure}
    \hfill
    \begin{subfigure}[b]{0.31\textwidth}
        \includegraphics[width=\linewidth]{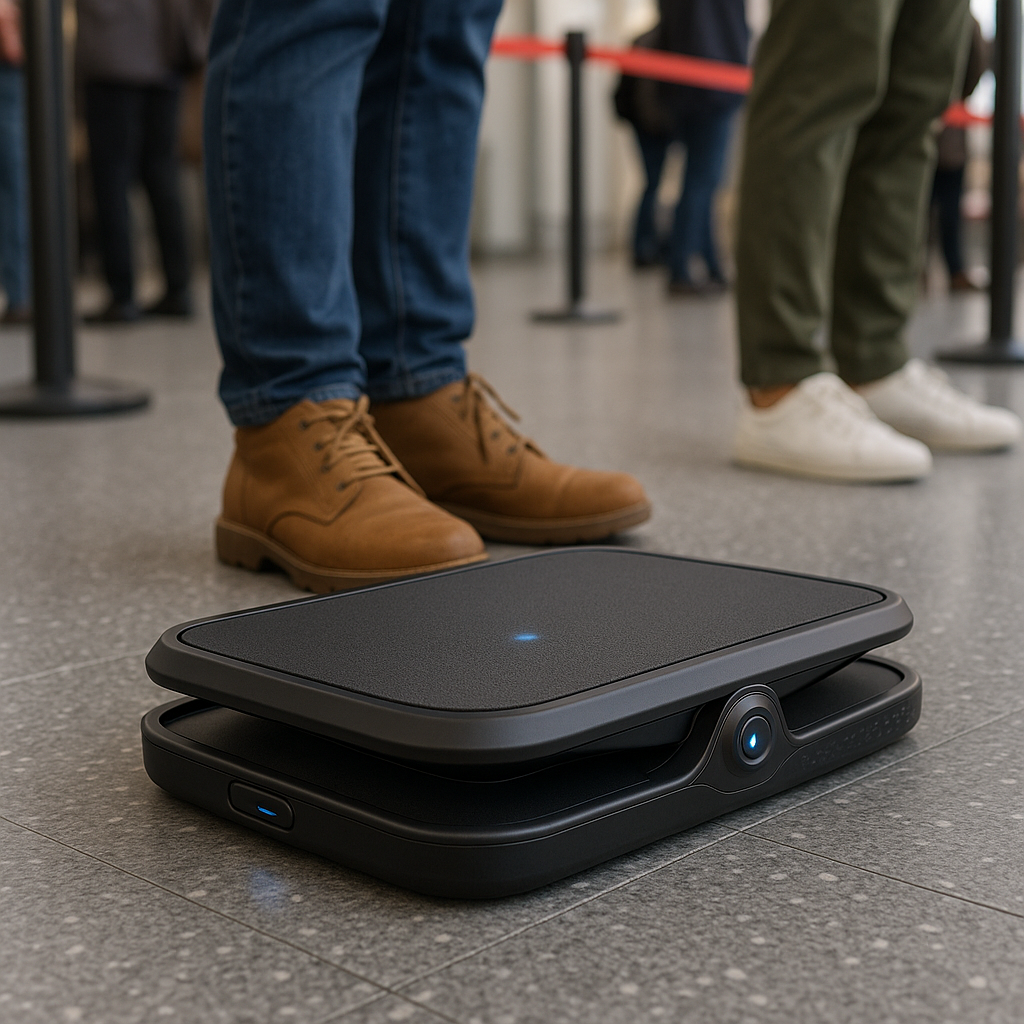}
        \caption{PS4: Self-Adjusting Posture Base}
    \end{subfigure}
    \vspace{0.2cm}

    \begin{subfigure}[b]{0.31\textwidth}
        \includegraphics[width=\linewidth]{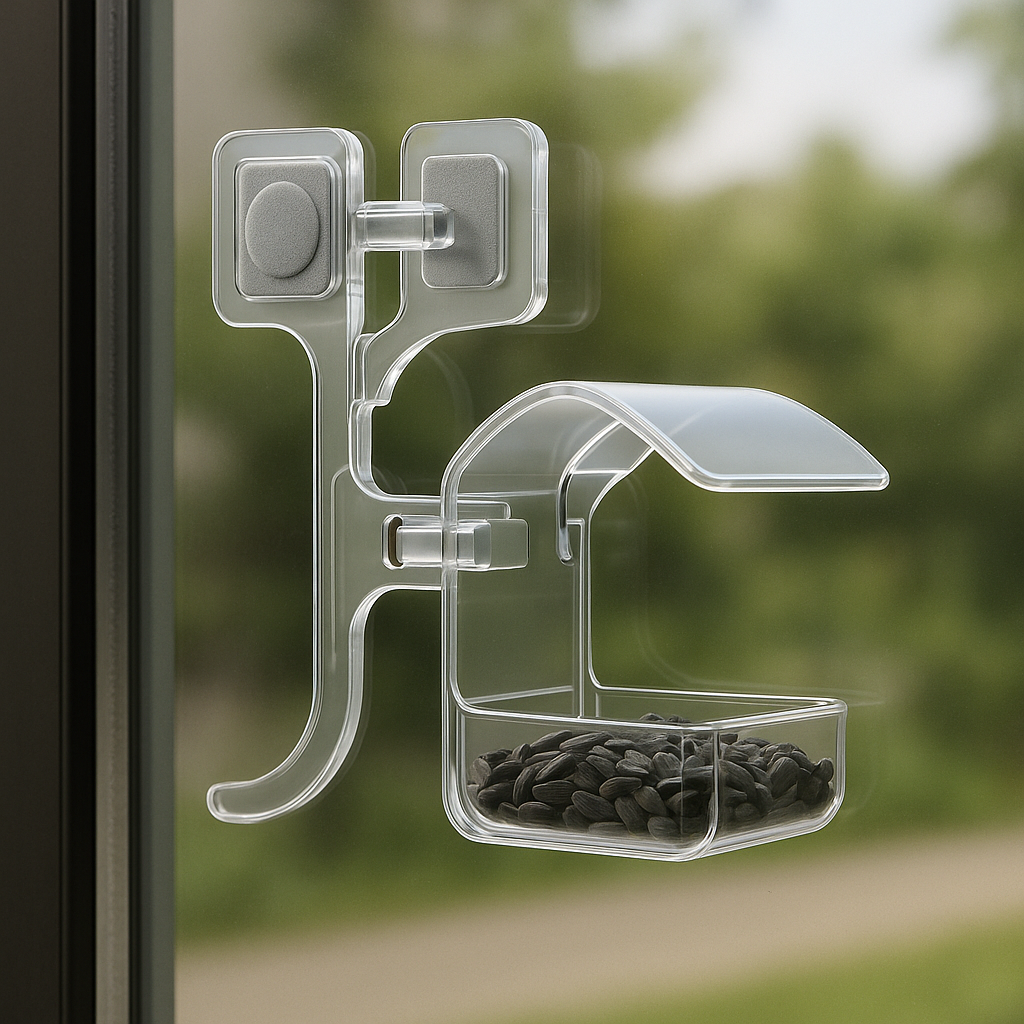}
        \caption{PS5: Stick-On Window Feeder}
    \end{subfigure}
    \hfill
    \begin{subfigure}[b]{0.31\textwidth}
        \includegraphics[width=\linewidth]{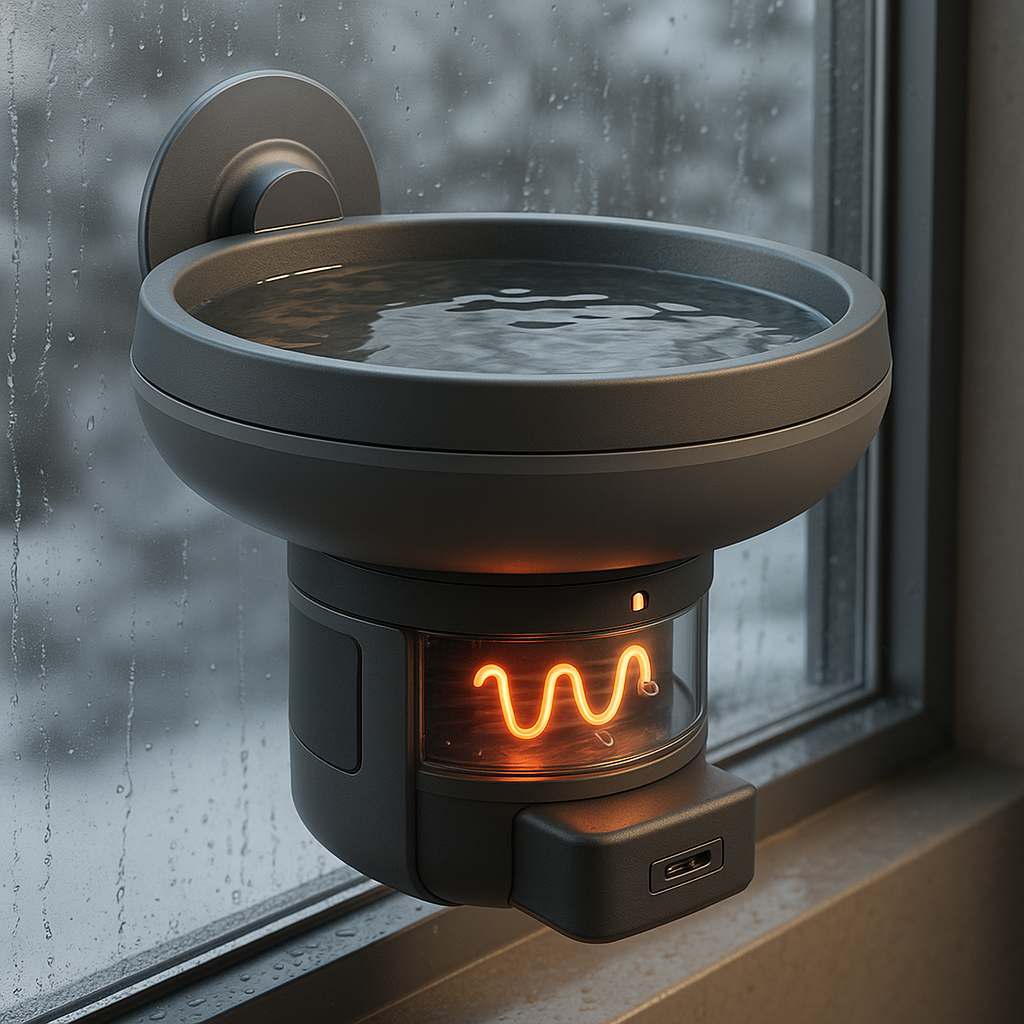}
        \caption{PS5: Winter Warm Water Feature}
    \end{subfigure}
    \hfill
    \begin{subfigure}[b]{0.31\textwidth}
        \includegraphics[width=\linewidth]{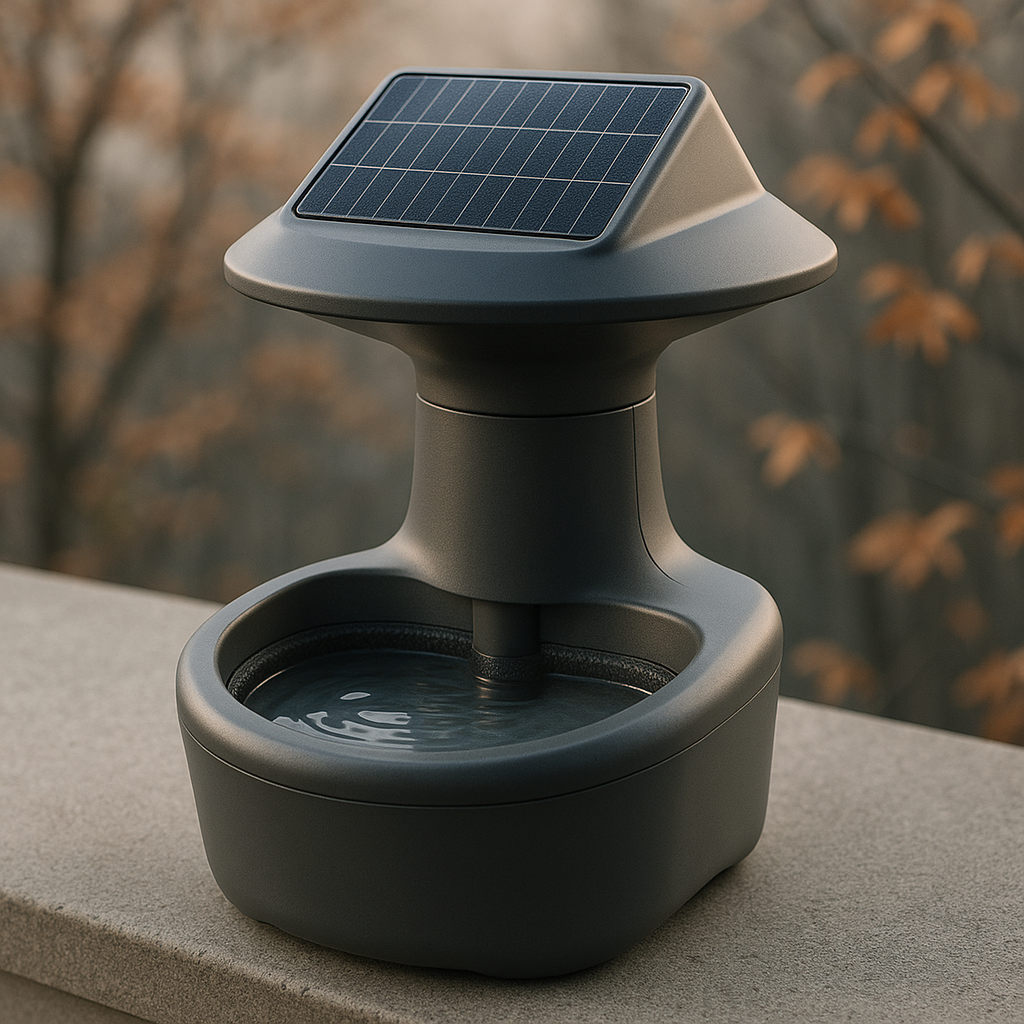}
        \caption{PS5: Frost-Proof Water Source}
    \end{subfigure}
    \vspace{0.2cm}

    \begin{subfigure}[b]{0.31\textwidth}
        \includegraphics[width=\linewidth]{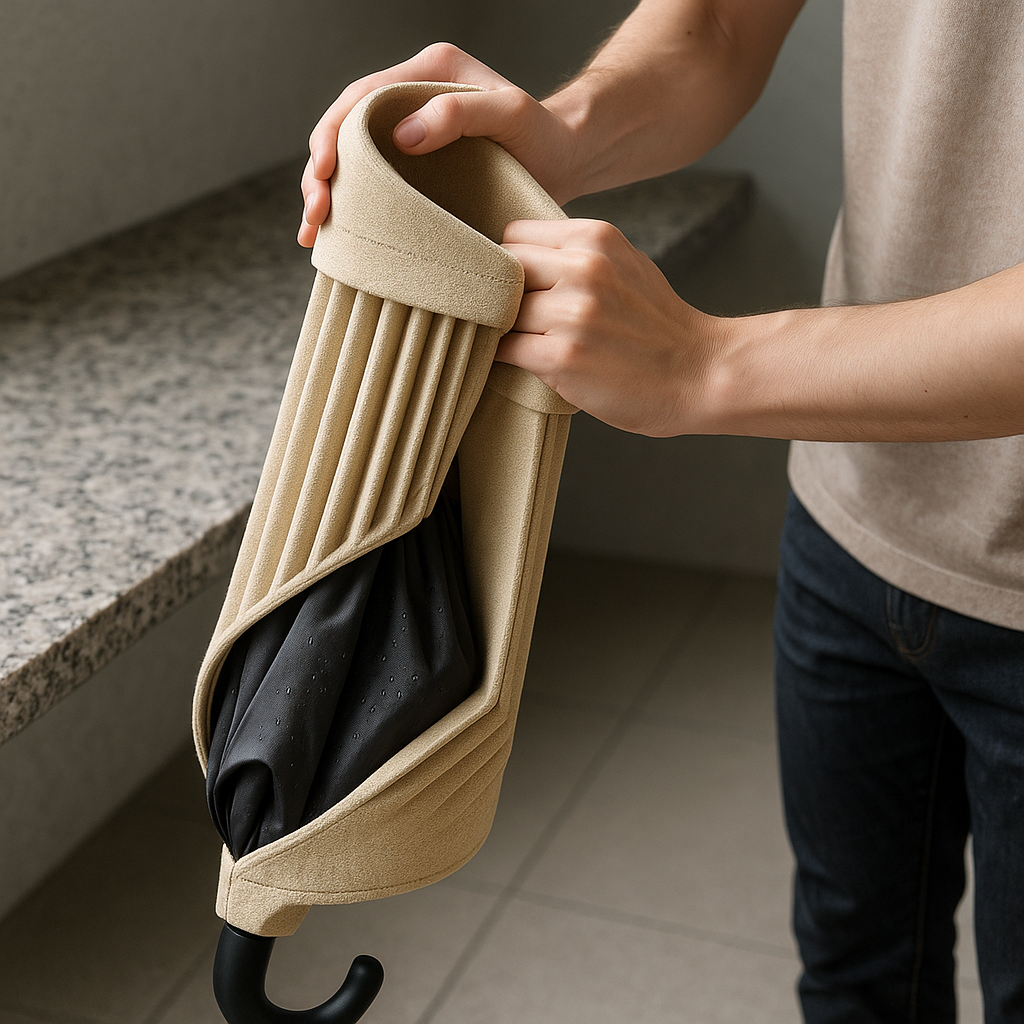}
        \caption{PS6: Accordion Squeeze Wrap}
    \end{subfigure}
    \hfill
    \begin{subfigure}[b]{0.31\textwidth}
        \includegraphics[width=\linewidth]{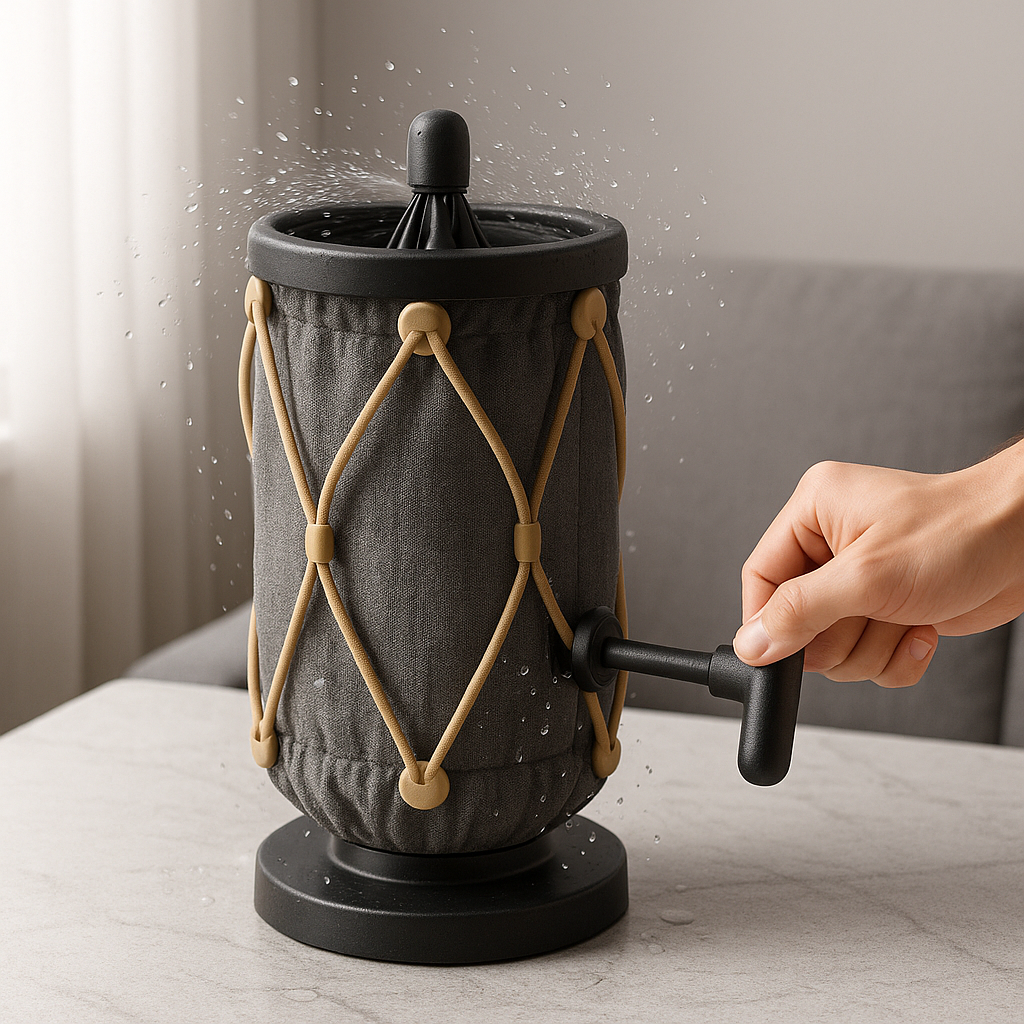}
        \caption{PS6: Centrifugal Stretch Spinner}
    \end{subfigure}
    \hfill
    \begin{subfigure}[b]{0.31\textwidth}
        \includegraphics[width=\linewidth]{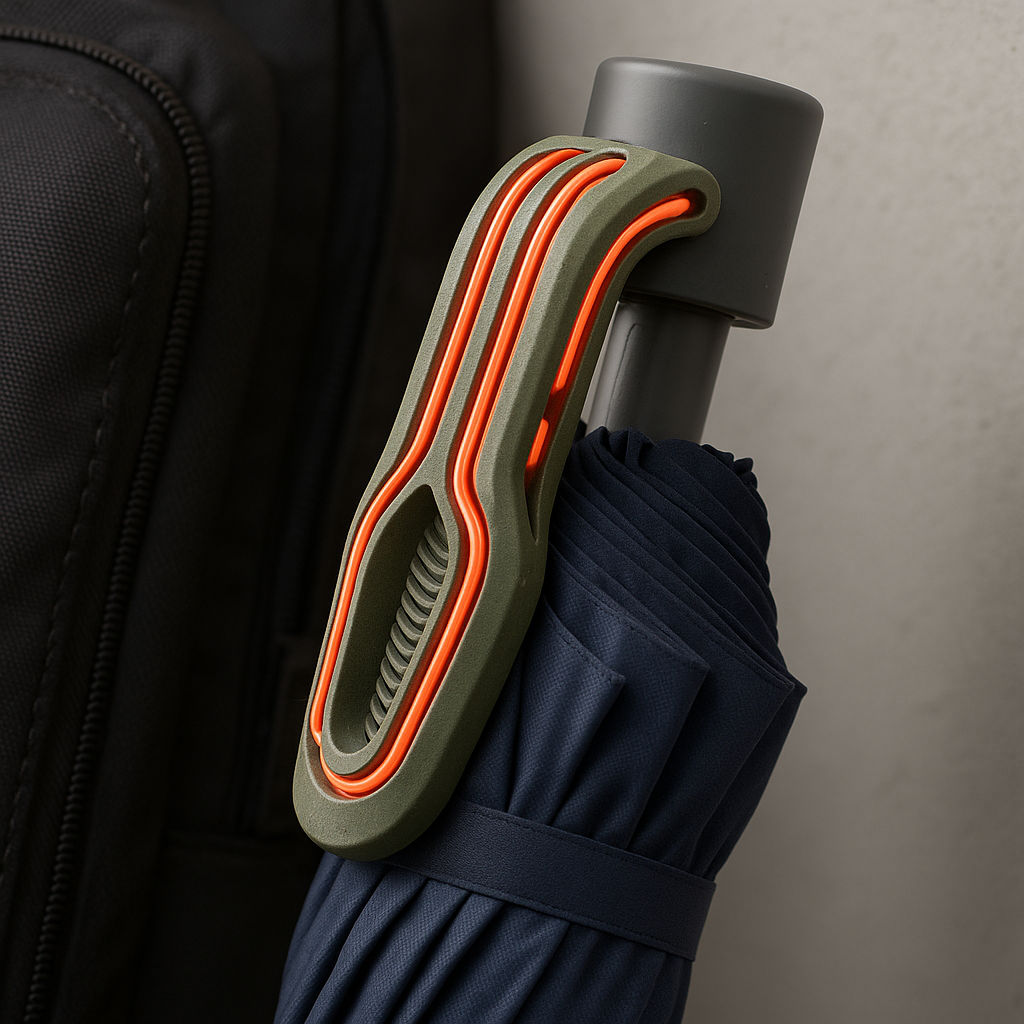}
        \caption{PS6: Passive Heat Circulation Latch}
    \end{subfigure}

    \caption{Concept renderings generated by the Leo agent for PS4–PS6. These visualizations represent the culmination of the progressive ideation pipeline, where structured PFIC concepts are translated into photorealistic forms to provoke final designer reflections.}
    \label{fig:leo_concept_renders_ps4_ps6}
\end{figure*}


The final stage of the MIDAS pipeline, managed by the \textit{Leo} agent, serves as the critical bridge between abstract semantic representations and tangible conceptualization. By translating the high-fidelity PFIC (Principle, Features, Implementation, Characteristics) data into photorealistic renderings as shown in Figure\ref{fig:leo_concept_renders_ps1_ps3} and Figure\ref{fig:leo_concept_renders_ps4_ps6}, the system operationalizes the Thought Provocation System (TPS) philosophy. These visualizations do not merely represent a final output but act as potent \textit{visual cues} that trigger a designer’s lateral thinking and reflective processes.

The ability of MIDAS to provide quick, high-quality renderings within a 20-minute workflow enables designers to evaluate the aesthetic and functional viability of diverse, non-overlapping ideas almost instantaneously. This rapid visual feedback loop encourages designers to revisit earlier stages of the pipeline with renewed clarity, ensuring that the human remains the central, active participant in a truly participatory, active and collaborative framework.

\section{Discussion}
\label{sec:discussion}

The results presented in Section \ref{sec:midas_demonstration} provide a compelling empirical basis for re-evaluating the role of Artificial Intelligence in the creative design process. By moving from a monolithic, single-shot generation paradigm to a distributed, agentic, and progressive workflow, the MIDAS framework addresses the deep-seated limitations of "idea abundance" that have plagued earlier computational ideation tools. This section synthesizes our findings to address the core research questions, exploring the theoretical and practical implications of agentic human-AI co-creation.

\subsection{Deconstructing the 'Novelty Illusion' through Agentic Specialization}
\label{sec:discussion_novelty_illusion}

A primary finding of this research is the stark contrast between the "apparent" novelty of single-shot LLM outputs and the "validated" novelty achieved through the MIDAS pipeline. The MIDAS framework successfully deconstructs the novelty illusion through \textit{Agentic Specialization}. By decomposing the ideation act into distinct roles, the system moves beyond simple interpolation. The \textit{Forge-Explorer} agent, for instance, is specifically parameterized to bypass high-probability tokens, while the \textit{Mint} and \textit{Scout} agents facilitate "ideas from ideas" through combinatorial synthesis. The "noise" clusters observed in Figure \ref{fig:midas_multi_cluster_plots} are its greatest success; they indicate that the system has successfully pushed the boundaries of the conceptual space so far that no two ideas share significant semantic overlap. This confirms that the MIDAS pipeline achieves a level of \textit{local novelty} and \textit{diversity} that is mathematically and creatively superior to traditional single-spurt generation.

\subsection{The Temporal Advantage: Operationalizing Progressive Cognition}
\label{sec:discussion_temporal}

Our research highlights a fundamental, yet often overlooked, dimension of creativity: \textit{Temporality}. We argued that human ideation is a path-dependent process where the $(n+1)^{th}$ idea is a reflective evolution of the $n$ ideas preceding it. Existing AI systems, by generating hundreds of ideas in a single instance, violate this temporal logic, leading to "conceptual stagnation" where the model essentially repeats itself within a single context window.

The MIDAS framework operationalizes this temporal order through its progressive pipeline. The \textit{Navigator} agent does not generate ideas in a vacuum; it builds upon the deconstructed Action-Object pairs from \textit{Mint} and the feasibility filters of \textit{Scout}. This incremental journey, from fuzzy problem statements to structured ideas, and then to deconstructed components before finally synthesizing into concepts, mimics the meta-cognitive "loops" of a human designer. The experimental results across PS1--PS6 (Figures \ref{fig:results_ps1_ps2}--\ref{fig:results_ps5_ps6}) confirm that this "progressive provocation" consistently yields ideas that are further away from the initial cluster than any single-shot spurt could achieve. This suggests that the key to breaking AI design fixation lies not in "larger" models, but in "slower," more structured, and temporally-aware interactions.

\subsection{Restoring Designer Agency: Reflections on the 'PAC' Model}
\label{sec:discussion_pac_reflections}

Perhaps the most significant qualitative outcome of this study is the success of the \textit{'PAC' (Participatory, Active, Collaborative)} model. In current AI-aided design literature, there is a palpable sense of "AI resistance" among professionals who feel that generative tools devalue their expertise. By relegating humans to the role of passive filters of machine-generated "slop," these tools foster disengagement and a loss of creative ownership.

The MIDAS framework reverses this power dynamic. The inclusion of the \textit{Muse} agent as the first generative step ensures that the designer’s "creative spark" is the foundation of the process. In our case demonstrations, designers reported a high sense of ownership precisely because their ideas were treated as equals to the machine’s contributions within the `Idea Vault`. The \textit{Sentinel} and \textit{Director} agents further reinforced this by allowing designers to prune and polish the final concept portfolio. This collaborative fluency demonstrates that AI is most powerful not when it replaces humans, but when it acts as a \textit{Thought Provocation System (TPS)}, a cognitive scaffold that nudges humans towards unexplored territories while respecting their role as the ultimate arbiter of value and relevance.

\subsection{Bridging the Gap: From Semantic Fluency to Global Innovation}
\label{sec:discussion_global_novelty}

A critical weakness in prevailing AI ideation research is the lack of "real-world grounding." An idea that is novel within the narrow context of a chat window may be trivial when compared to the vast history of human invention. MIDAS bridges this gap through the \textit{Librarian} and \textit{Challenger} agents. By integrating real-time web search and patent benchmarking into the pipeline, the system transitions from "linguistic creativity" to "technical innovation."

The filtering metrics in Table \ref{tab:agent_filtering_metrics} reveal a significant "funnel effect." While \textit{Forge} might generate 70+ ideas, the \textit{Challenger} agent typically filters this down to a handful of globally novel candidates. This rigorous global novelty check ensures that the designer's time is spent only on concepts that have the potential to be truly state-of-the-art. This grounding is essential for a "reputed conference" context, where innovation is measured against the prior art. The ability of MIDAS to consistently produce such high-quality, globally novel outputs within a 20-minute window marks a significant leap in the efficiency and efficacy of Computer-Aided Innovation (CAI).

\subsection{Implications for Design Practice and Pedagogy}
\label{sec:discussion_implications}

The success of MIDAS has profound implications for both design practice and education. For industry practitioners, MIDAS offers a way to overcome "evaluation overload." Instead of wasting hours sorting through variant-heavy brainstorming logs, designers can arrive at a diverse portfolio of 10 to 15 high-quality, feasible, and novel concepts in under half an hour. This efficiency enables more frequent and in-depth iterations in the early stages of the design process.

In a pedagogical context, the structured nature of the agents provides a "cognitive roadmap" for students. By interacting with agents like \textit{Scribe} (for problem definition) or \textit{Mint} (for divergence), novice designers learn the \textit{methodology} of good design while simultaneously benefiting from the AI's vast knowledge base. MIDAS thus functions as both a tool for production and a tool for learning, reinforcing the systematic design approaches taught in top-tier design schools \cite{Beitz1996}.

\subsection{Limitations}
\label{sec:discussion_limitations}

While the preliminary results from our pilot study are highly encouraging, we must acknowledge the limitations of this initial work. 
\begin{itemize}
    \item The current system relies on text and image-based LLMs; future iterations could benefit from integrating multi-modal agents capable of reasoning over 3D geometry or functional physics simulations. 
    \item Furthermore, while the 20-minute session time is a strength, longitudinal studies are needed to understand how sustained use of such a system affects a designer's long-term creative habits. 
    \item The study was conducted with a small, homogeneous group of six novice designers (n=6). While this provided rich qualitative feedback, the findings cannot be generalized to the broader population of expert designers or industry professionals without further validation. 
    \item The design problems, while complex, were well-defined product-design scenarios; the framework's performance on more ambiguous, systemic, or "wicked" problems remains to be tested.
    \item The confounding variables, such as the novelty of the MIDAS system itself or potential learning effects resulting from the study's sequential design, warrant a more controlled and larger-scale investigation.
\end{itemize}

Nevertheless, the results across all six problems provide a robust validation of the agentic approach. We have demonstrated that by treating ideation as a progressive, multi-agent, and participatory journey, we can transcend the "stochastic median" of current AI and unlock a new era of truly innovative human-AI co-creation.

\subsection{Directions for Future Research}
\label{sec:future_work}

This research opens several promising avenues for future inquiry, building directly upon the limitations and insights from this study.

\begin{enumerate}
    \item \textbf{Large-Scale Validation:} The immediate next step is to conduct larger, more diverse validation studies. This includes recruiting participants with varying levels of expertise (e.g., novice students vs. 20-year industry veterans) and from different domains (e.g., engineering, service design, UX/UI) to assess the framework's broader utility and adaptability.
    
    \item \textbf{Refinement of Agentic Capabilities:} Future work will focus on enhancing the specialization of individual agents. For instance, the `Librarian` agent could be augmented to parse multi-modal patent databases that include CAD files and technical diagrams, not just text. The `Scout` agent could be integrated with engineering knowledge graphs or basic physics simulations to provide far more robust feasibility scoring, moving beyond LLM-based intuition.
        
    \item \textbf{Deeper Human-AI Interaction Research:} The 'PAC' model merits a dedicated HCI investigation. Longitudinal studies could explore how a designer's creative process and cognitive skills evolve over time with sustained use of the MIDAS system. We are also interested in exploring adaptive agency, where the system could dynamically adjust the level of "provocation" or support based on the designer's detected expertise or cognitive state.
    
    \item \textbf{Hybrid Intelligence Architectures:} While MIDAS currently employs a compound LLM approach, we will explore integrating non-LLM AI, such as symbolic reasoning engines or evolutionary algorithms, for specific agents. A hybrid-intelligence `Scout` (for feasibility) or `Sentinel` (for relevance) could offer a powerful combination of generative fluency and logical rigor.
    
    \item \textbf{From Visualization to Fabrication:} The output of the current pipeline ends with `Leo` (image generation). A significant future goal is to extend this pipeline, developing a `Fabricator` agent that can translate the final PFIC concepts into preliminary CAD models, 3D-printable files, or bills of materials, thereby bridging the gap between concept and tangible prototype.
\end{enumerate}

In summary, MIDAS represents a foundational step towards a new generation of agentic, progressive, and truly collaborative design tools. The path forward involves not only refining this technology but also deepening our understanding of the new creative partnerships with the humans.

\section{Conclusion}
\label{sec:conclusion}
This paper addresses the inadequacy of single-spurt, monolithic AI systems for the complex, multi-stage process of creative ideation. This work demonstrated that moving from a single AI "partner" to a distributed "team" of specialized, collaborative agents is a more effective model for human-AI co-creativity in generating a truly novel cohort of ideas. The principal contribution of this paper is the MIDAS framework and system. MIDAS operationalizes the 'PAC' model for human-AI partnership, respecting the designer's agency and fostering shared ownership. Agentic pipeline emulates meta-cognition via 'CG/CA' workflow, progressively refining ideas to innovative concepts. Therefore, this research demonstrates a viable path towards a synergistic partnership where the machine manages complex analysis and broad exploration, allowing the human to focus on providing context, exercising judgement, and directing the creative journey. Thus, in conclusion, MIDAS represents a new generation of agentic, progressive, active, participatory, and truly collaborative design paradigm for human-AI co-creation.




\bibliographystyle{ACM-Reference-Format}
\bibliography{bibliography}


\appendix



\end{document}